\documentclass{article} 
\usepackage{collas2024_conference,times}
\usepackage{easyReview}
\usepackage{subcaption}
\usepackage{float}

\newcommand{\minisection}[1]{\vspace{0.04in} \noindent {\bf #1}\ \ }

\usepackage{amsmath,amsfonts,bm}

\def\eqref#1{equation~\ref{#1}}

\def\1{\bm{1}}

\DeclareMathAlphabet{\mathsfit}{\encodingdefault}{\sfdefault}{m}{sl}
\SetMathAlphabet{\mathsfit}{bold}{\encodingdefault}{\sfdefault}{bx}{n}

\usepackage{hyperref}
\hypersetup{
    colorlinks=true,
    linkcolor=red,
    filecolor=magenta,
    urlcolor=blue,
    citecolor=purple,
    pdftitle={Overleaf Example},
    pdfpagemode=FullScreen,
}

\title{An Empirical Analysis of Forgetting in Pre-trained Models with Incremental Low-Rank Updates}

\author{
Albin Soutif--Cormerais\textsuperscript{1}\textsuperscript{2}, Simone Magistri\textsuperscript{3}, Joost van de Weijer\textsuperscript{1}\textsuperscript{2}, Andrew D. Bagdanov\textsuperscript{3}
\vspace{2mm} \\
\textsuperscript{1} Computer Vision Center, Spain \\
\textsuperscript{2} Department of Computer Science, Universitat Autònoma de Barcelona, Spain \\
\textsuperscript{3} Department of Information Engineering, University of Florence, Italy \\
\textsuperscript{1}\textsuperscript{2} \texttt{\{albin, joost\}@cvc.uab.es} \\
\textsuperscript{3} \texttt{\{simone.magistri, andrew.bagdanov\}@unifi.it} \\
}

\preprintcopy 

\begin{document}


\newcommand{\figloraftshort}[3]{
\begin{figure}[H]
  \centering
  \begin{subfigure}[b]{0.4\textwidth}
    \includegraphics[trim=#1, clip, width=#2\linewidth]{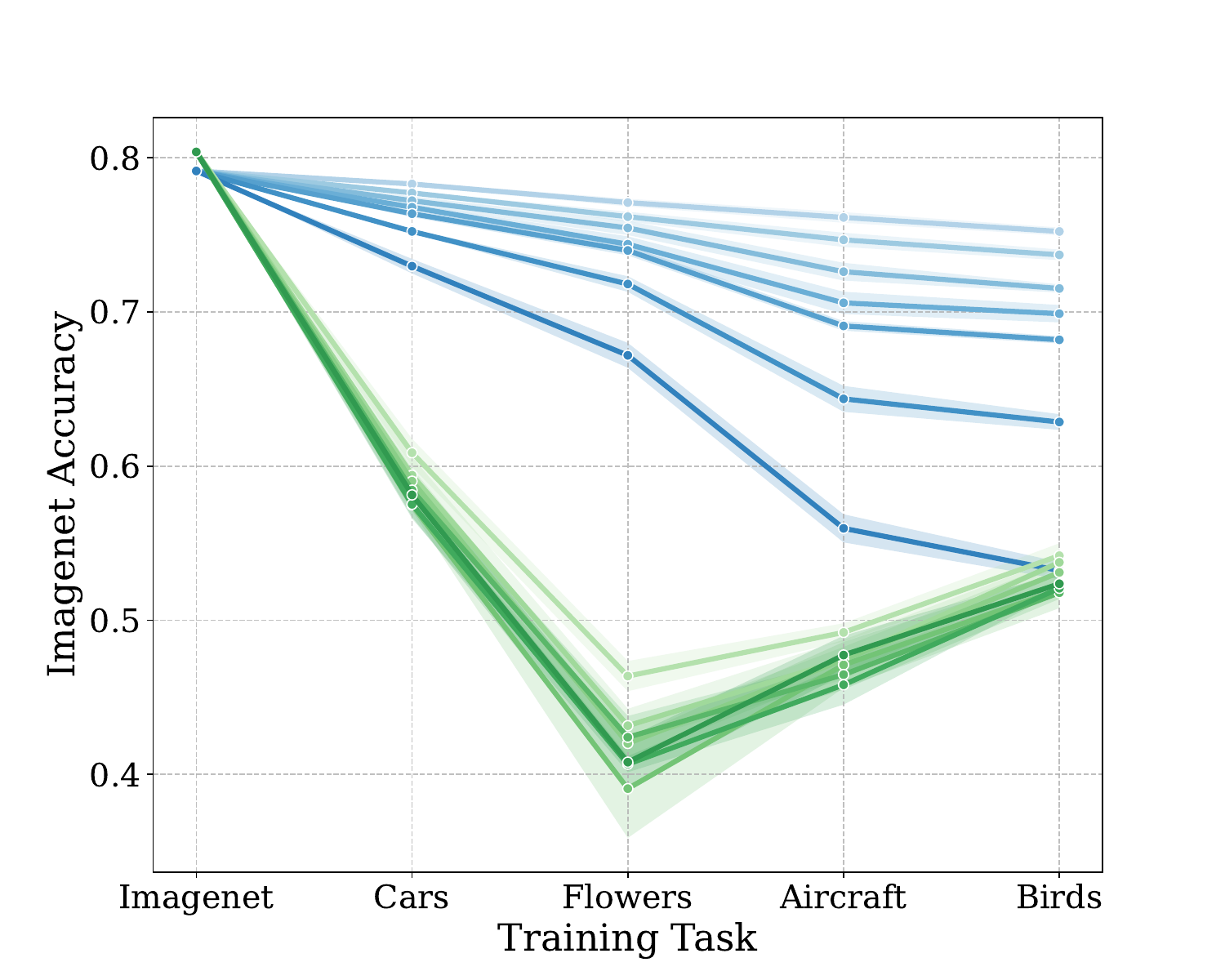}
  \end{subfigure}%
  \begin{subfigure}[b]{0.4\textwidth}
    \includegraphics[trim=#1, clip, width=#2\linewidth]{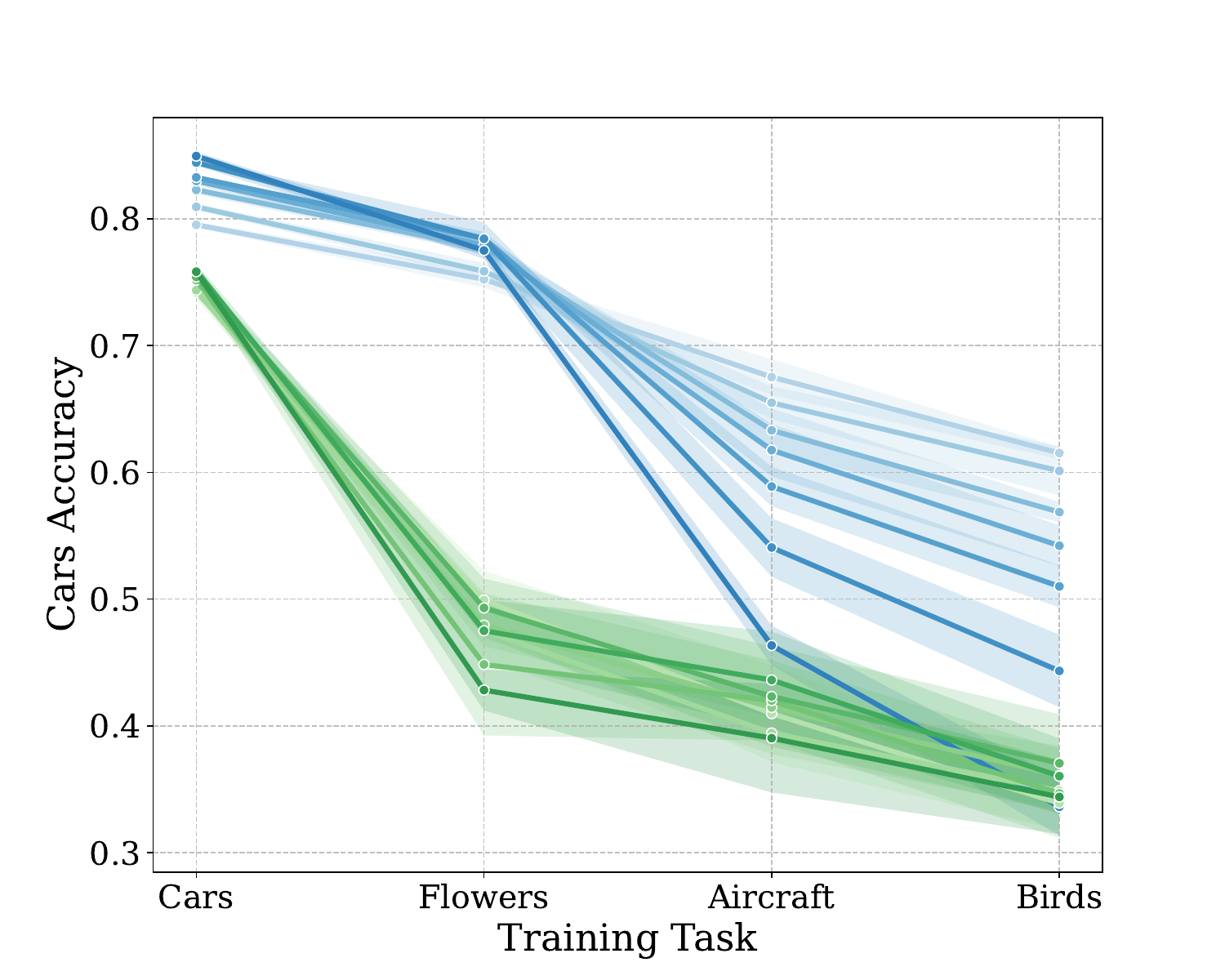}
  \end{subfigure}%
  \begin{subfigure}[b]{0.2\textwidth}
    \includegraphics[trim=400 0 0 0, clip, width=0.85\linewidth]{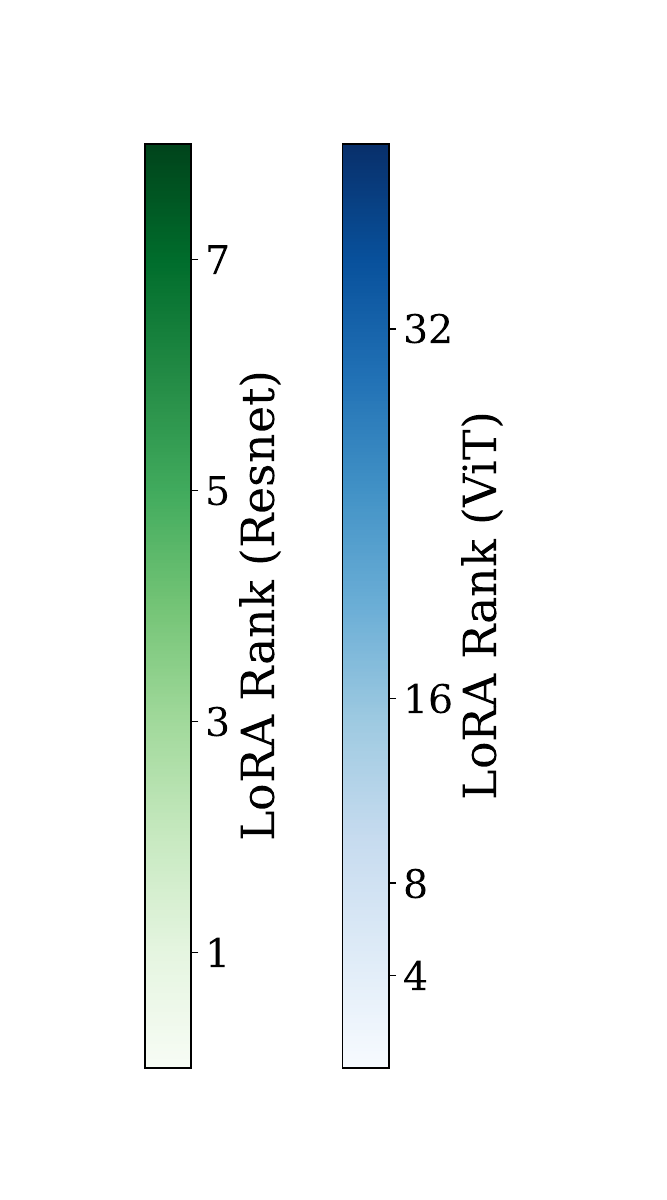}
  \end{subfigure}
  \caption{#3}
  \label{fig:lora_ft}
\end{figure}
}

\newcommand{\figloralwf}[3]{
\begin{figure}[H]
  \centering
  \begin{subfigure}[b]{0.4\textwidth}
    \includegraphics[trim=#1, clip, width=#2\linewidth]{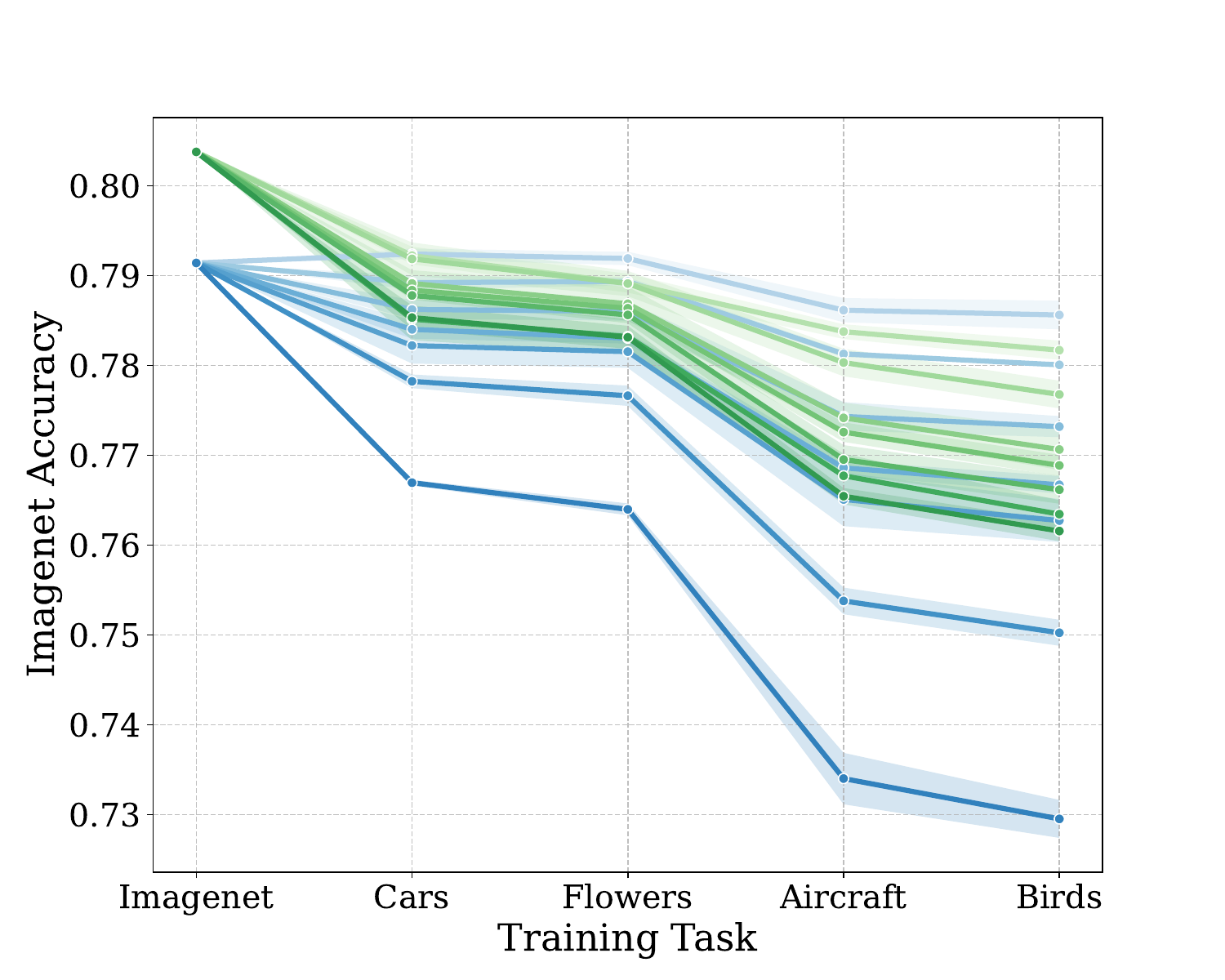}
  \end{subfigure}%
  \begin{subfigure}[b]{0.4\textwidth}
    \includegraphics[trim=#1, clip, width=#2\linewidth]{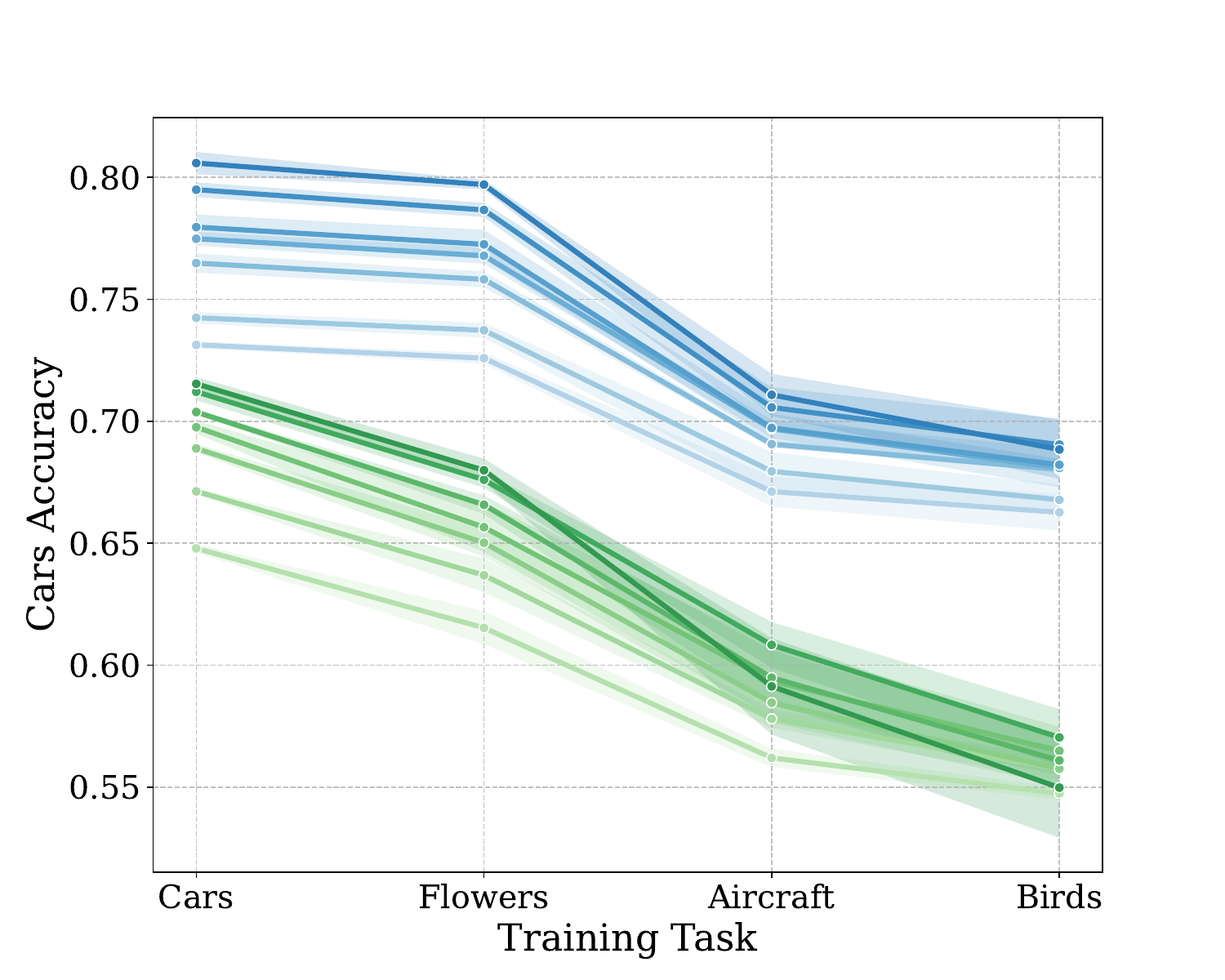}
  \end{subfigure}%
  \begin{subfigure}[b]{0.2\textwidth}
    \includegraphics[trim=400 0 0 0, clip, width=0.85\linewidth]{figures/legend_nofull.pdf}
  \end{subfigure}
  \caption{#3}
  \label{fig:lora_lwf}
\end{figure}
}

\newcommand{\figbarplot}[3]{
\begin{figure}[H]
  \centering
  \begin{subfigure}[b]{0.5\textwidth}
    \includegraphics[trim=#1, clip, width=#2\linewidth]{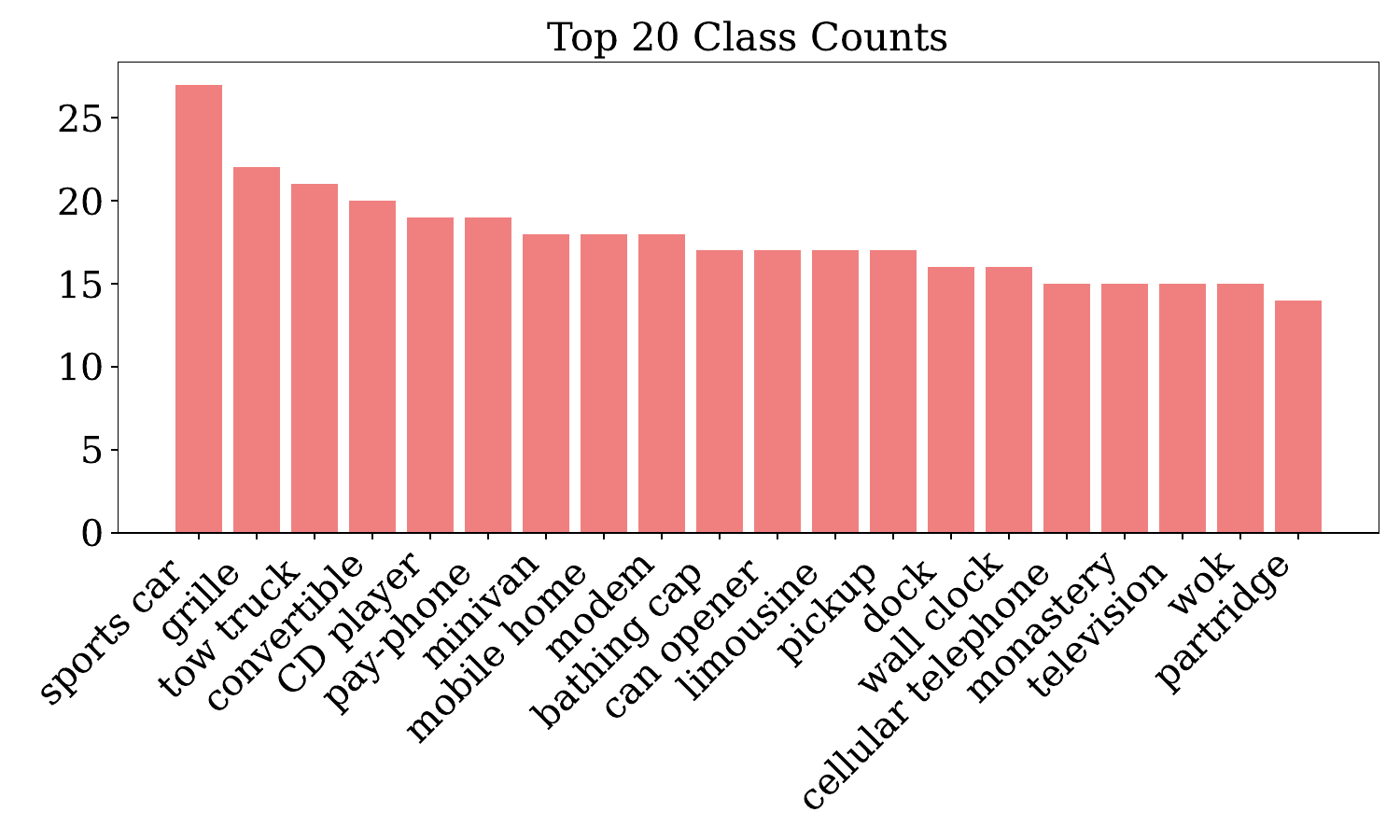}
  \end{subfigure}%
  \begin{subfigure}[b]{0.5\textwidth}
    \includegraphics[trim=#1, clip, width=#2\linewidth]{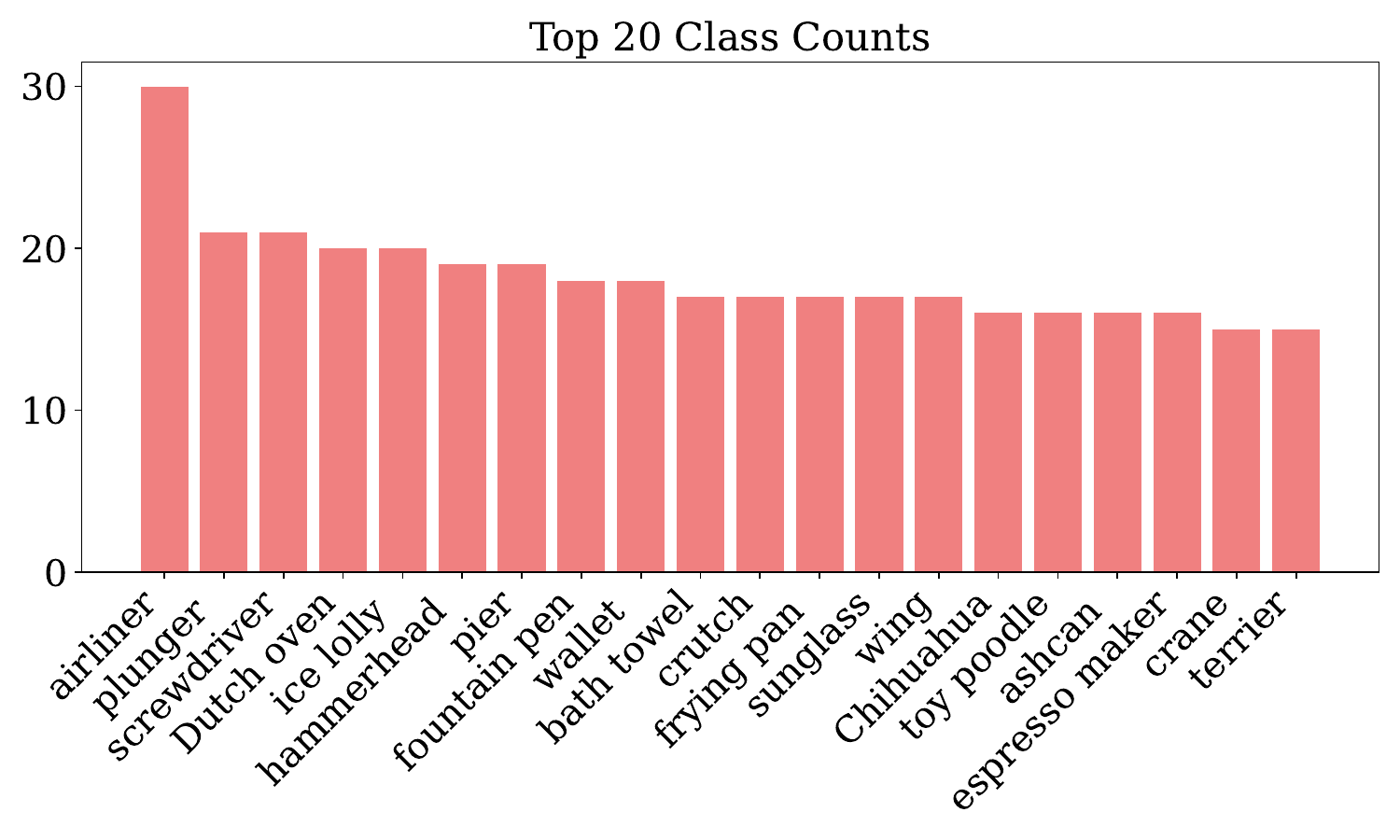}
  \end{subfigure}
  \caption{#3}
  \label{fig:forget_hist}
\end{figure}
}

\newcommand{\figloraftlong}[3]{
\begin{figure}[H]
  \centering
  \begin{subfigure}[b]{0.4\textwidth}
    \includegraphics[trim=#1, clip, width=#2\linewidth]{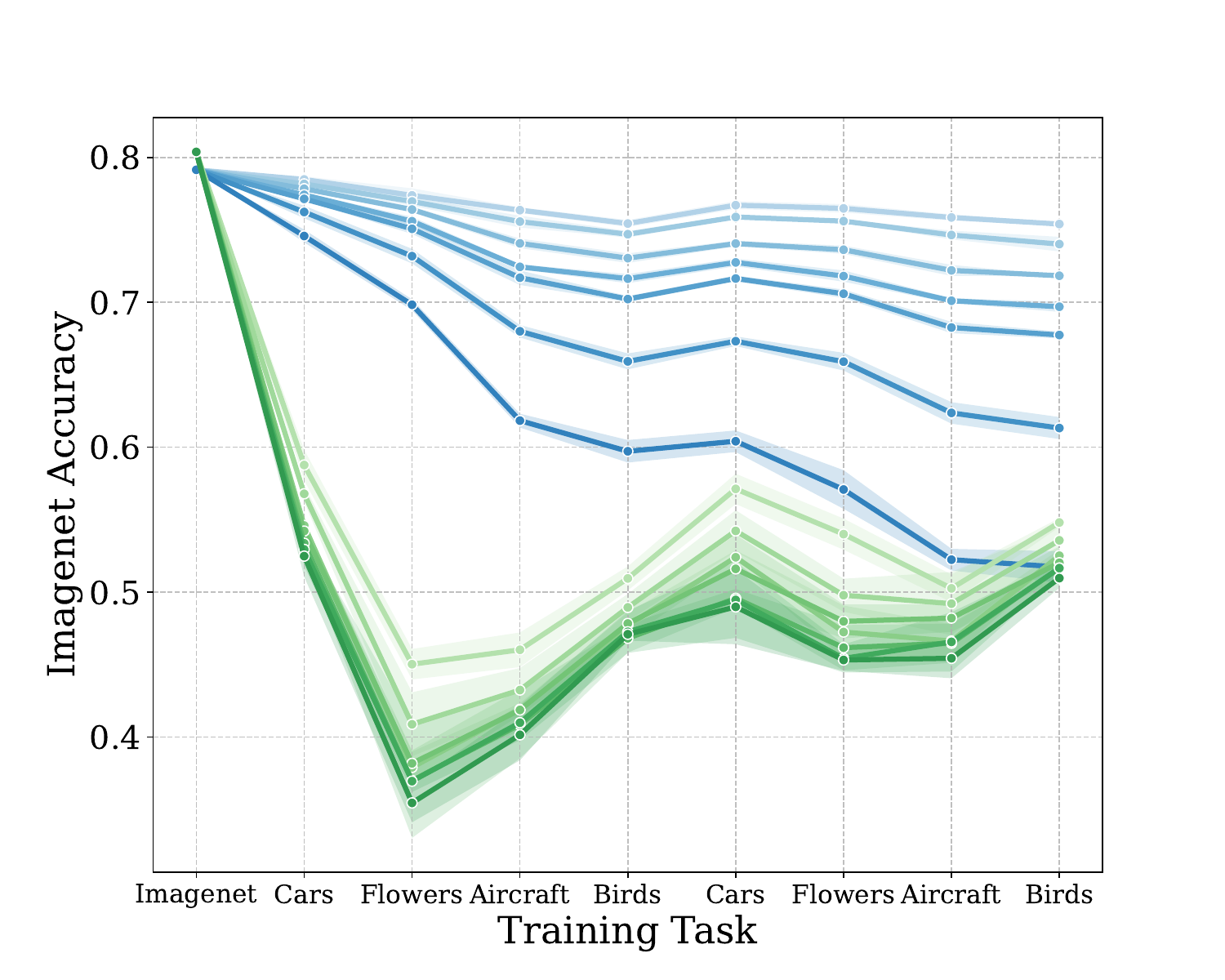}
  \end{subfigure}%
  \begin{subfigure}[b]{0.4\textwidth}
    \includegraphics[trim=#1, clip, width=#2\linewidth]{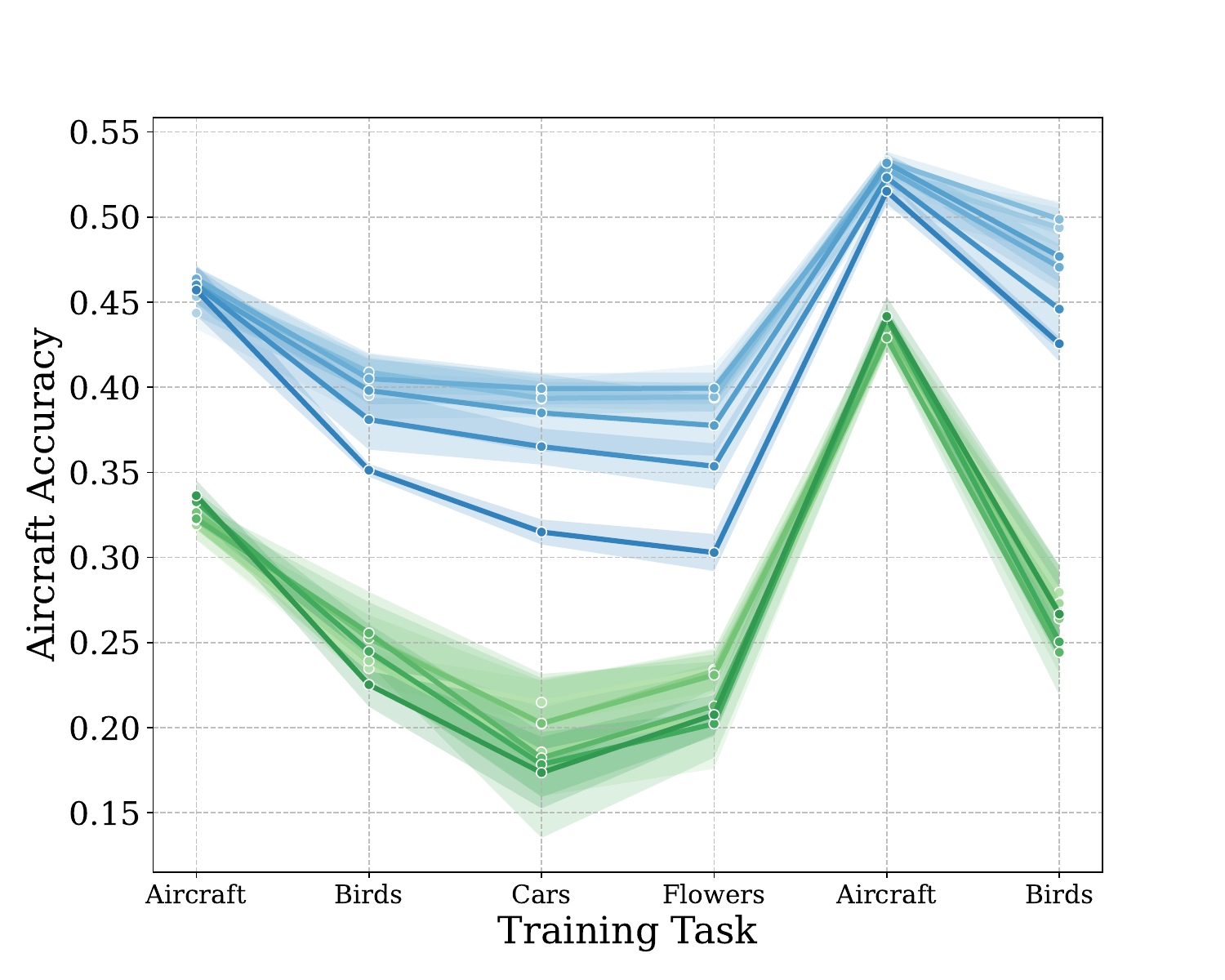}
  \end{subfigure}%
  \begin{subfigure}[b]{0.2\textwidth}
    \includegraphics[trim=400 0 0 0, clip, width=0.85\linewidth]{figures/legend_nofull.pdf}
  \end{subfigure}
  \caption{#3}
  \label{fig:lora_ft_long}
\end{figure}
}

\newcommand{\figloraftlongtwenty}[3]{
\begin{figure}[H]
  \centering
  \begin{subfigure}[b]{0.4\textwidth}
    \includegraphics[trim=#1, clip, width=#2\linewidth]{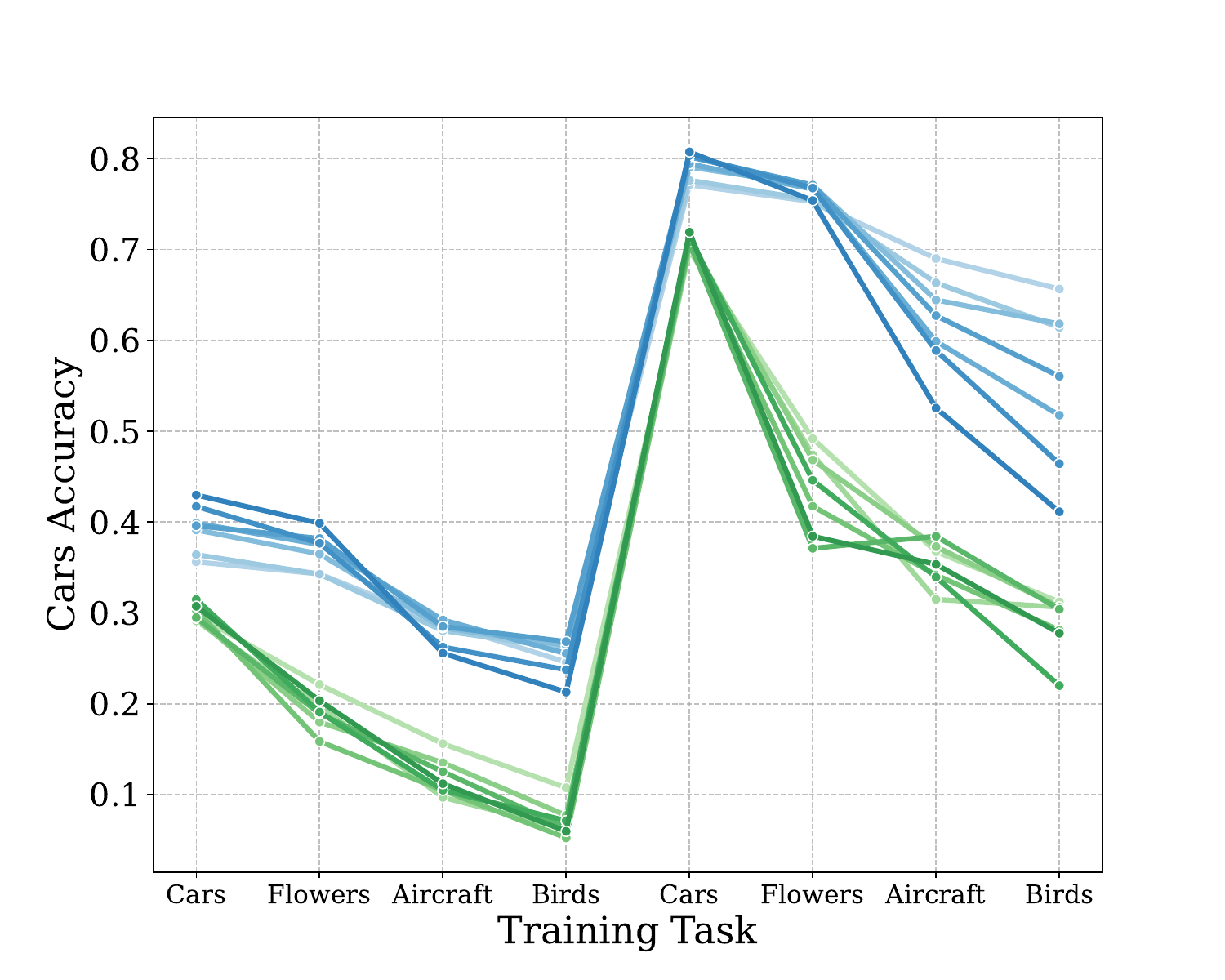}
  \end{subfigure}%
  \begin{subfigure}[b]{0.4\textwidth}
    \includegraphics[trim=#1, clip, width=#2\linewidth]{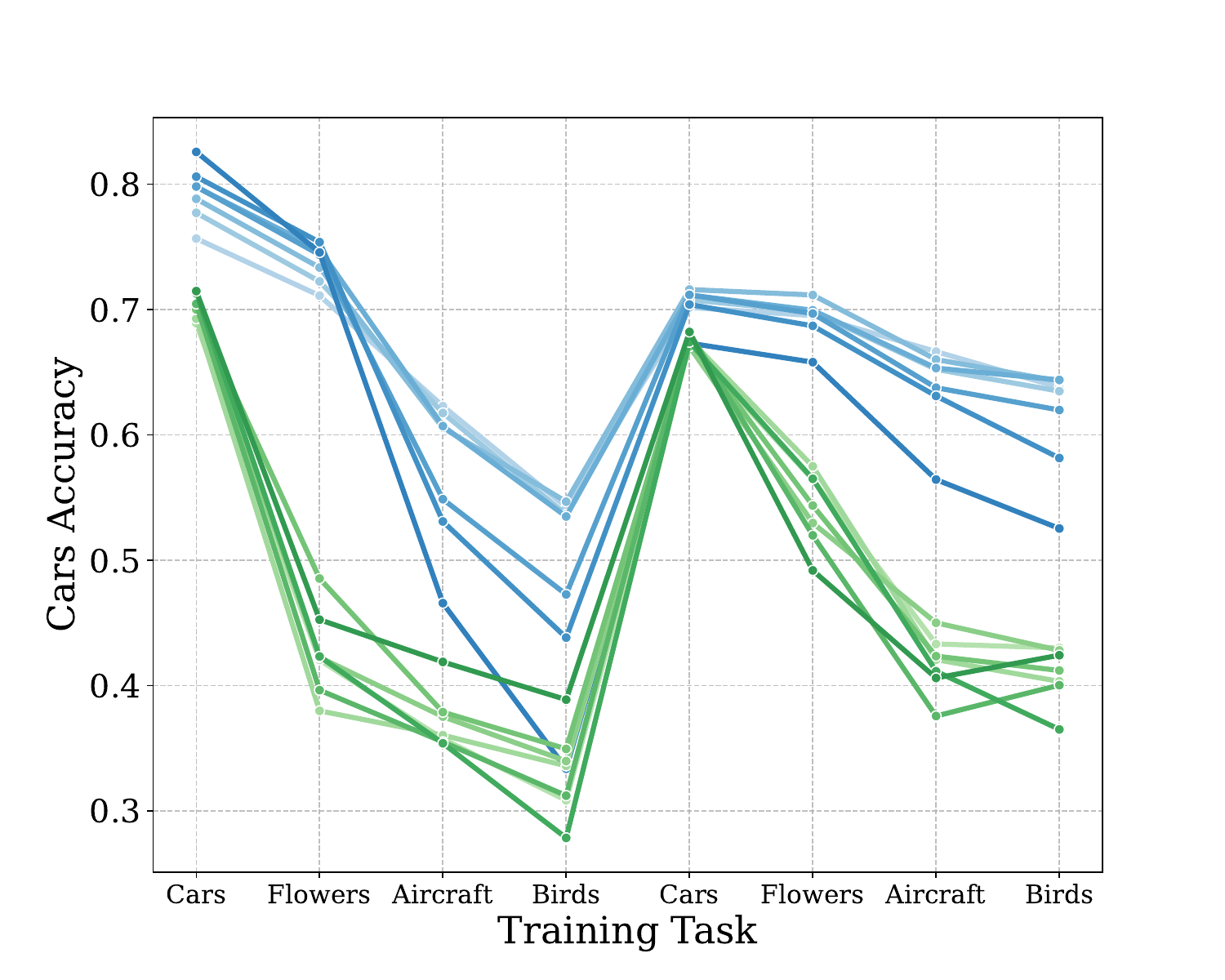}
  \end{subfigure}%
  \begin{subfigure}[b]{0.2\textwidth}
    \includegraphics[trim=400 0 0 0, clip, width=0.85\linewidth]{figures/legend_nofull.pdf}
  \end{subfigure}
  \caption{#3}
  \label{fig:lora_ft_long_8020}
\end{figure}
}

\newcommand{\figloraftavg}[3]{
\begin{figure}[H]
  \centering
  \begin{subfigure}[b]{0.4\textwidth}
    \includegraphics[trim=#1, clip, width=#2\linewidth]{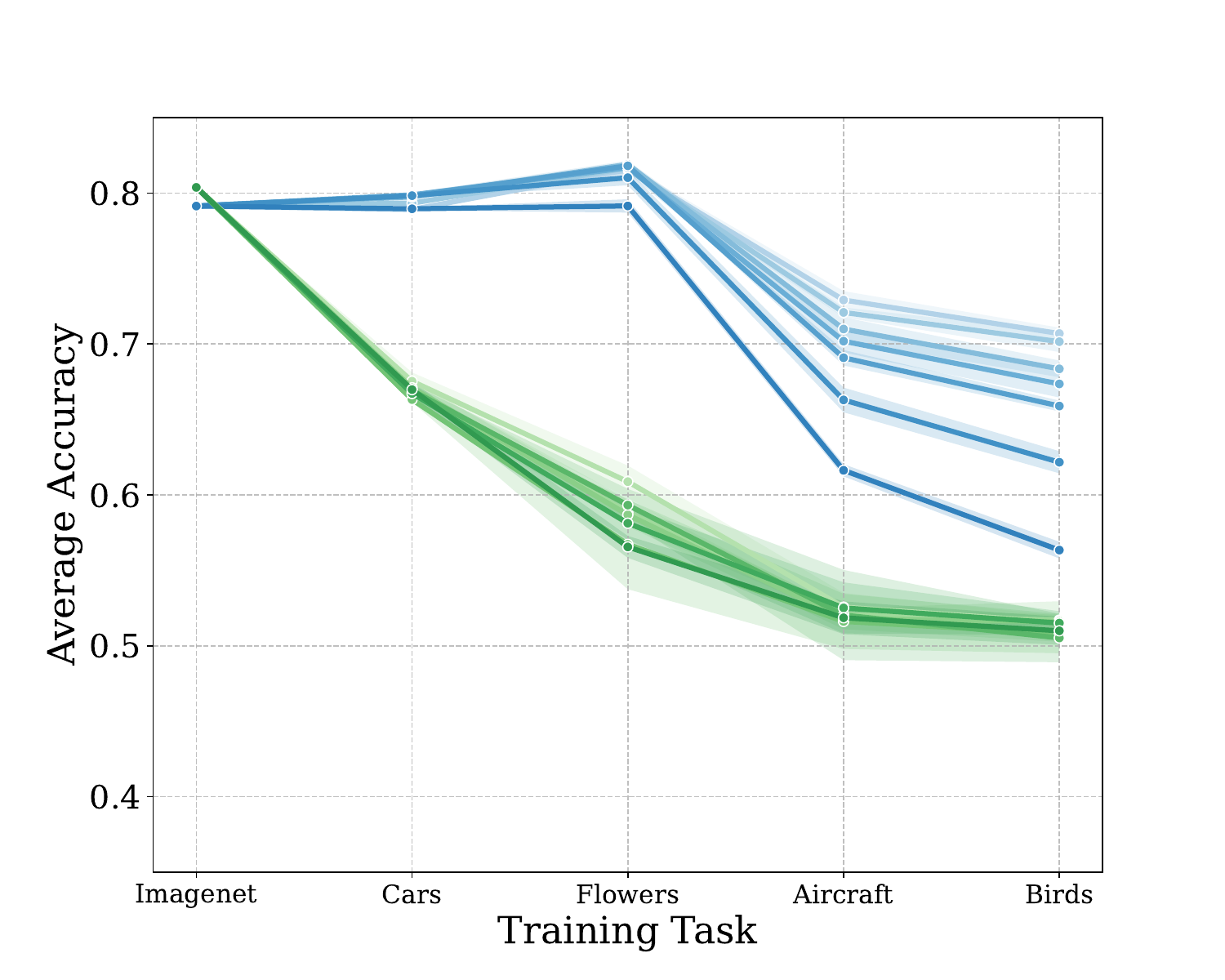}
  \end{subfigure}%
  \begin{subfigure}[b]{0.4\textwidth}
    \includegraphics[trim=#1, clip, width=#2\linewidth]{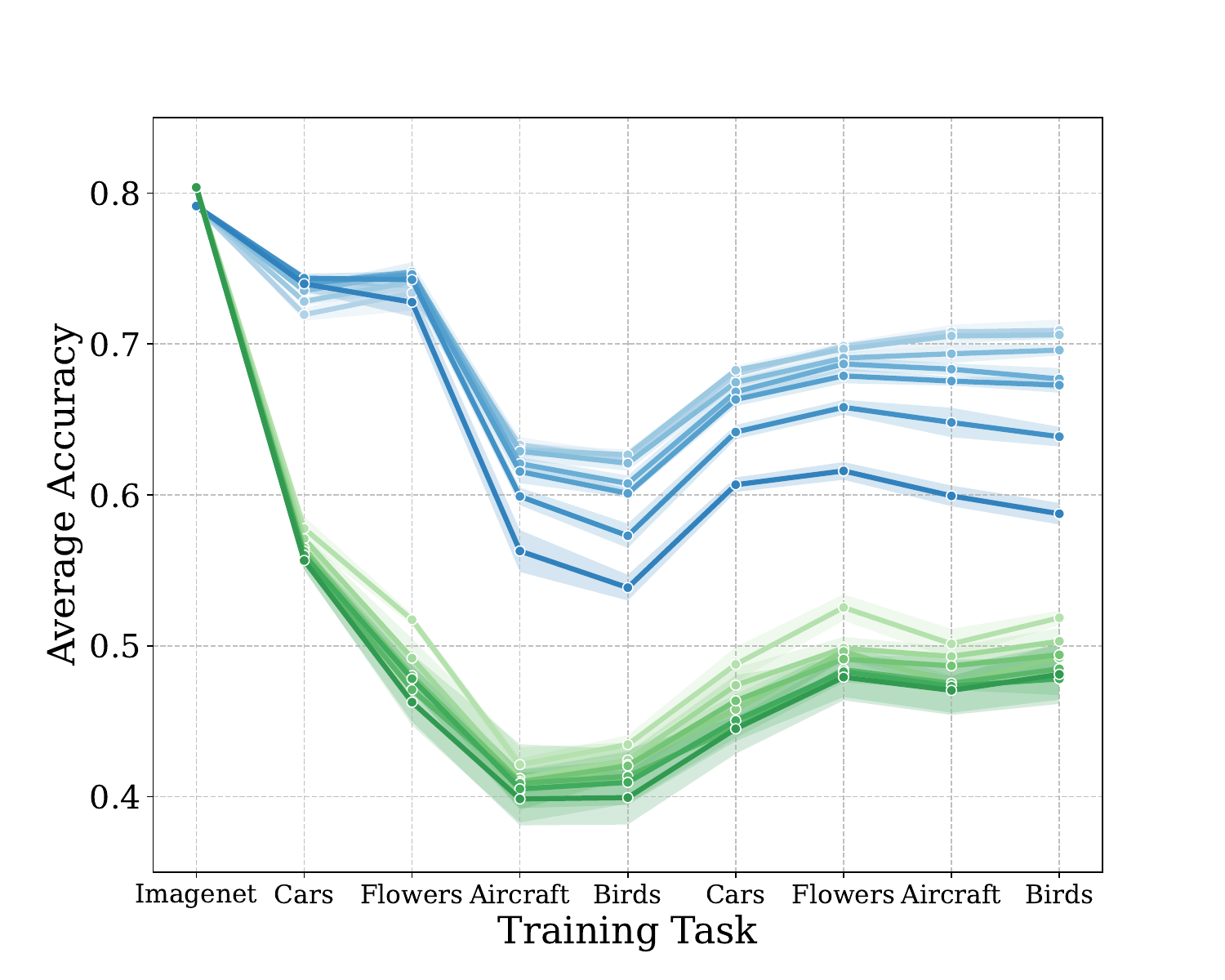}
  \end{subfigure}%
  \begin{subfigure}[b]{0.2\textwidth}
    \includegraphics[trim=400 0 0 0, clip, width=0.85\linewidth]{figures/legend_nofull.pdf}
  \end{subfigure}
  \caption{#3}
  \label{fig:lora_avg}
\end{figure}
}

\newcommand{\figkde}[3]{
\begin{figure}[H]
  \centering
  \begin{subfigure}[b]{0.5\textwidth}
    \includegraphics[trim=#1, clip, width=#2\linewidth]{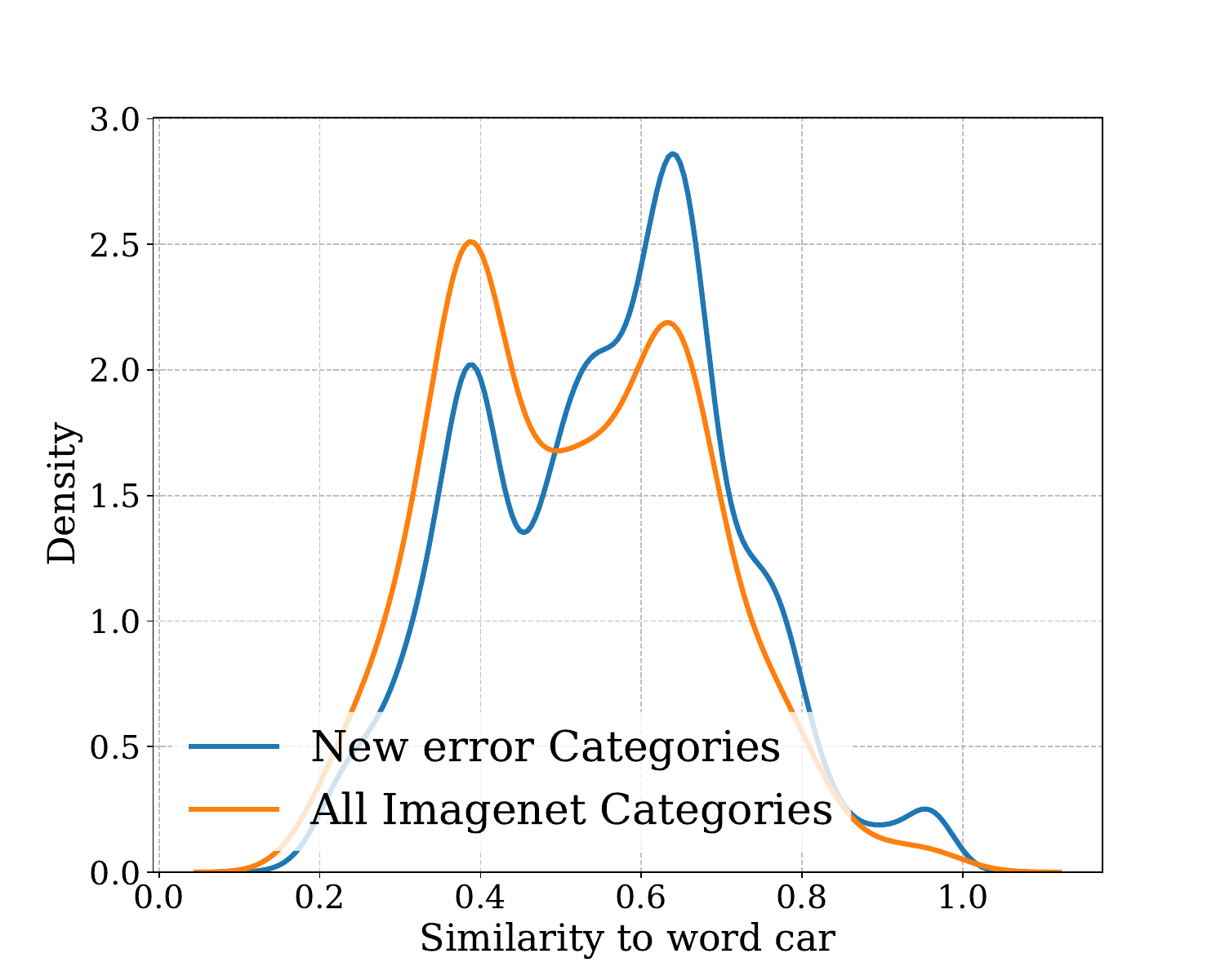}
  \end{subfigure}%
  \begin{subfigure}[b]{0.5\textwidth}
    \includegraphics[trim=#1, clip, width=#2\linewidth]{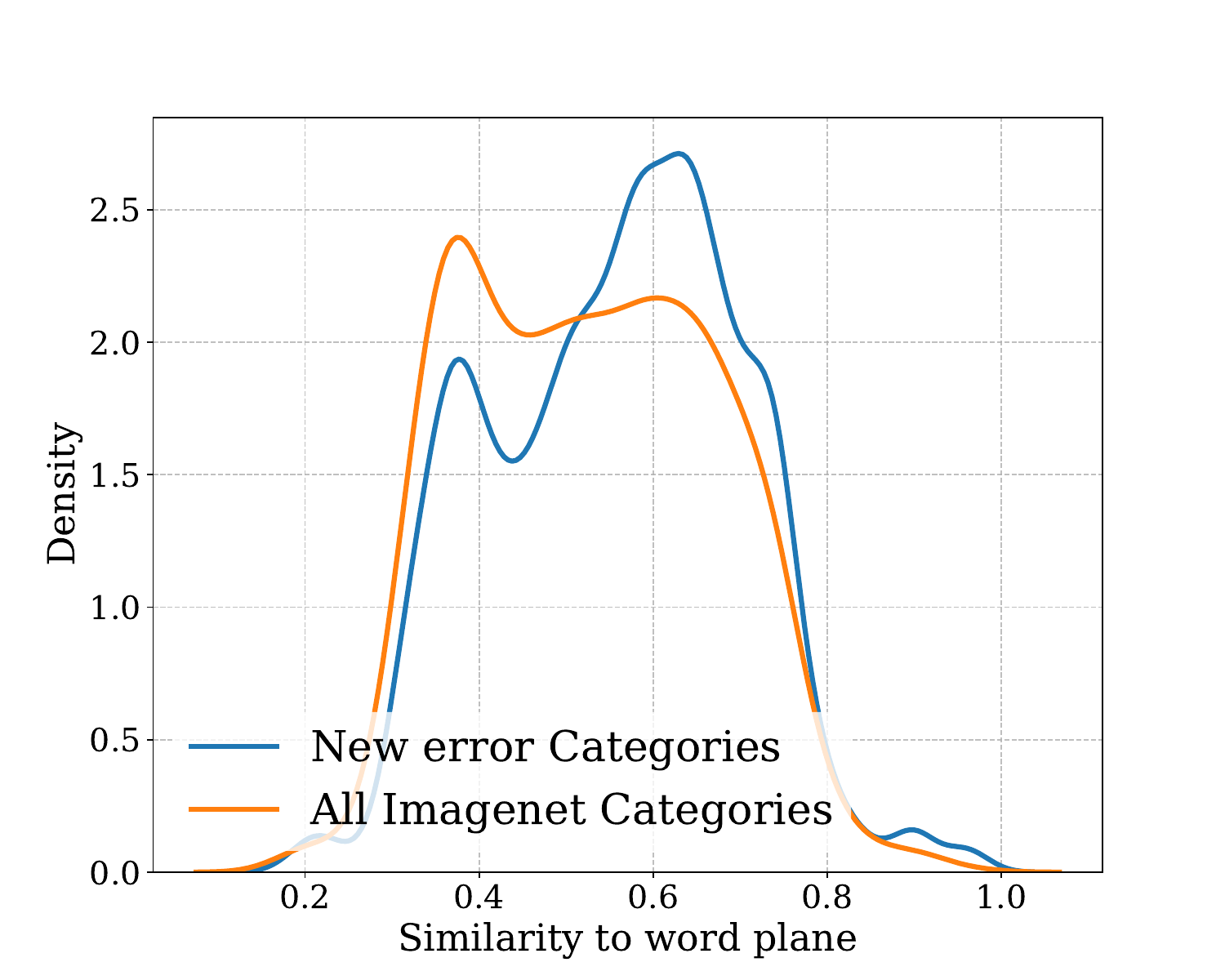}
  \end{subfigure}
  \caption{#3}
  \label{fig:forget_kde}
\end{figure}
}

\newcommand{\figkderesnet}[3]{
\begin{figure}[H]
  \centering
  \begin{subfigure}[b]{0.5\textwidth}
    \includegraphics[trim=#1, clip, width=#2\linewidth]{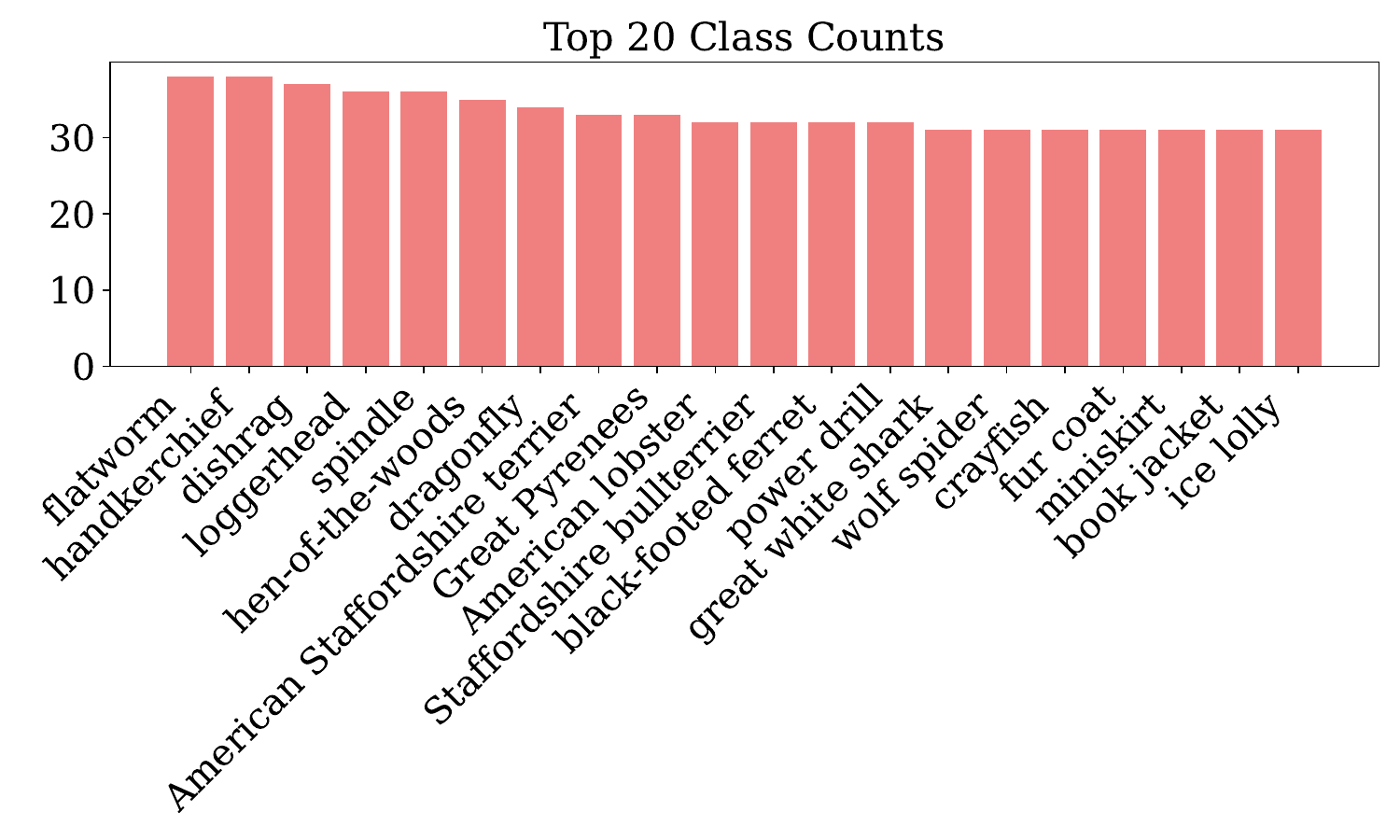}
  \end{subfigure}%
  \begin{subfigure}[b]{0.5\textwidth}
    \includegraphics[trim=#1, clip, width=#2\linewidth]{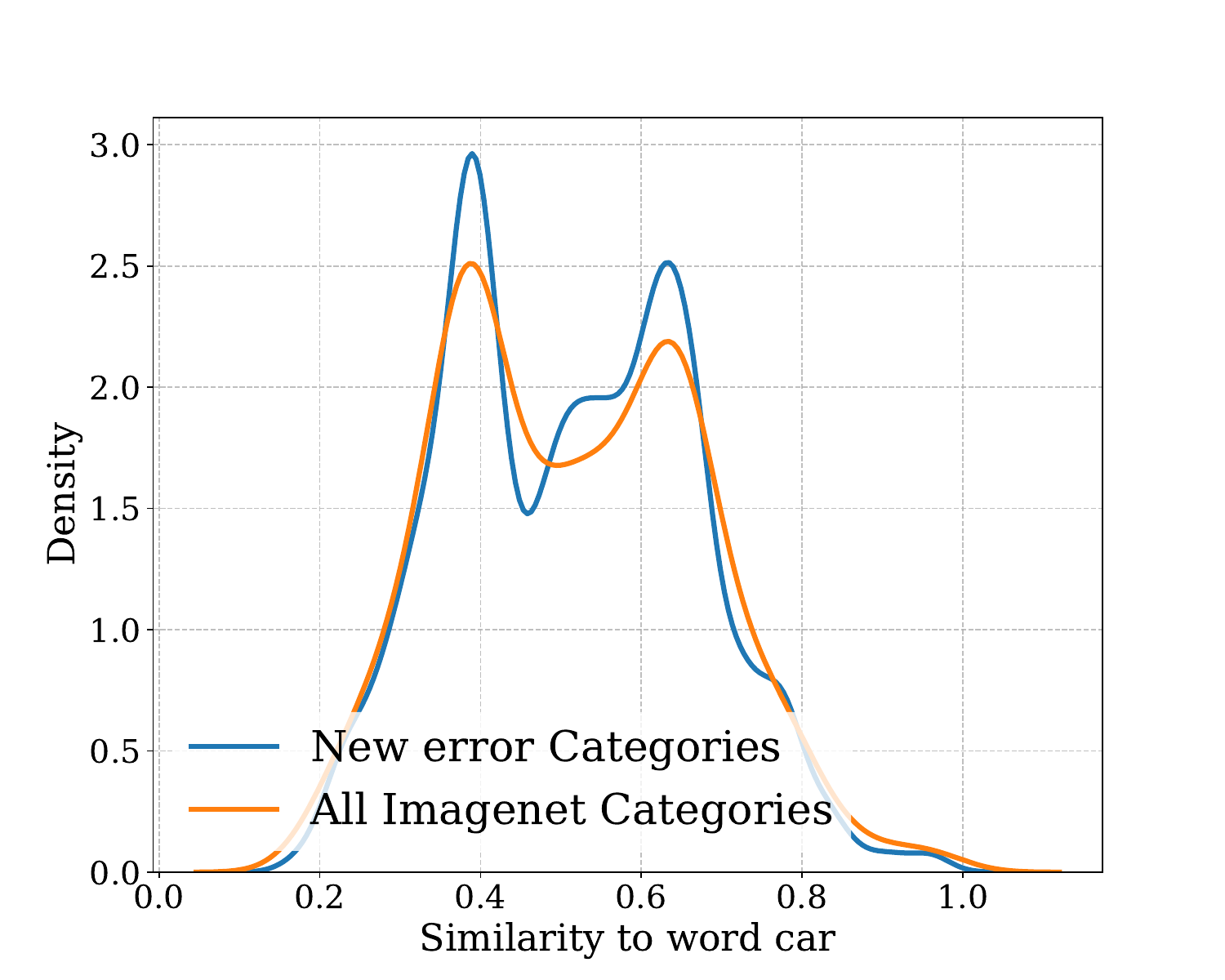}
  \end{subfigure}
  \caption{#3}
  \label{fig:context_resnet}
\end{figure}
}

\newcommand{\tableforgettingshort}[1]{
\begin{table}[H]
    \centering
    \begin{minipage}{0.5\textwidth}
        \centering
        \begin{tabular}{ccc}
            \hline
            \textbf{Rank} & \textbf{Average Accuracy} & \textbf{Average Forgetting} \\
            \hline
            1 & 69.9 & 9.2 \\
            4 & 66.8 & 14.3 \\
            8 & 68.0 & 13.6 \\
            16 & 62.2 & 21.0 \\
            32 & 55.9 & 28.9 \\
            \hline
        \end{tabular}
    \end{minipage}%
    \begin{minipage}{0.5\textwidth}
        \centering
        \begin{tabular}{ccc}
            \hline
            \textbf{Rank} & \textbf{Average Accuracy} & \textbf{Average Forgetting} \\
            \hline
            1 & 52.2 & 23.6 \\
            3 & 51.2 & 25.5 \\
            4  & 52.0 & 24.8 \\
            5 & 50.3 & 27.8 \\
            7 & 47.9 & 30.7 \\
            \hline
        \end{tabular}
    \end{minipage}
    \caption{#1}
    \label{tab:table_forgetting}
\end{table}
}

\newcommand{\tableforgettingshortnew}[1]{
\begin{table}[H]
    \centering
    \begin{minipage}{0.5\textwidth}
        \centering
        \begin{tabular}{ccc}
            \hline
            \textbf{Rank} & \textbf{Average Accuracy} & \textbf{Average Forgetting} \\
            \hline
            Head & 62.4  & 0 \\
            1 & 70.7 $\pm$ 0.4 & 8.4 $\pm$ 0.42 \\
            2 & 70.2 $\pm$ 0.7 & 9.7 $\pm$ 1.0 \\
            4 & 68.3 $\pm$ 0.5 & 12.6 $\pm$ 0.7 \\
            6 & 67.3 $\pm$ 0.9 & 14.2 $\pm$ 1.1 \\
            8 & 65.9 $\pm$ 0.3 & 16.2 $\pm$ 0.4 \\
            16 & 62.1 $\pm$ 0.7 & 20.8 $\pm$ 0.8 \\
            32 & 56.3 $\pm$ 0.5 & 27.9 $\pm$ 0.9 \\
            Full & 12.7 $\pm$ 0.3 & 63.3 $\pm$ 0.8 \\
            \hline
        \end{tabular}
    \end{minipage}%
    \begin{minipage}{0.5\textwidth}
        \centering
        \begin{tabular}{ccc}
            \hline
            \textbf{Rank} & \textbf{Average Accuracy} & \textbf{Average Forgetting} \\
            \hline
            Head & 60.0 & 0 \\
            1 & 51.6 $\pm$ 0.4 & 25.0 $\pm$ 0.5 \\
            2 & 51.7 $\pm$ 0.4 & 24.7 $\pm$ 0.8 \\
            3 & 51.6 $\pm$ 1.3 & 25.3 $\pm$ 1.6 \\
            4 & 50.8 $\pm$ 1.3 & 26.5 $\pm$ 1.4 \\
            5 & 50.5 $\pm$ 1.6 & 27.1 $\pm$ 2.0 \\
            6 & 51.5 $\pm$ 0.8 & 26.0 $\pm$ 1.2 \\
            7 & 50.9 $\pm$ 0.9 & 26.7 $\pm$ 1.3 \\
            Full & 41.8 $\pm$ 1.3 & 44.6 $\pm$ 2.1 \\
            \hline
        \end{tabular}
    \end{minipage}
    \caption{#1}
    \label{tab:table_forgetting}
\end{table}
}

\maketitle

\begin{abstract}
Broad, open source availability of large pretrained foundation models on the internet through platforms such as HuggingFace has taken the world of practical deep learning by storm. A classical pipeline for neural network training now typically consists of finetuning these pretrained network on a small target dataset instead of training from scratch. In the case of large models this can be done even on modest hardware using a low rank training technique known as Low-Rank Adaptation (LoRA). While Low Rank training has already been studied in the continual learning setting, existing works often consider storing the learned adapter along with the existing model but rarely attempt to modify the weights of the pretrained model by merging the LoRA with the existing weights after finishing the training of each task. In this article we investigate this setting and study the impact of LoRA rank on the forgetting of the pretraining foundation task and on the plasticity and forgetting of subsequent ones. We observe that this rank has an important impact on forgetting of both the pretraining and downstream tasks. We also observe that vision transformers finetuned in that way exhibit a sort of ``contextual'' forgetting, a behaviour that we do not observe for residual networks and that we believe has not been observed yet in previous continual learning works.
\end{abstract}

\section{Introduction}


The capability of machine learning to automatically learn from data and extract discriminative patters patterns has led to groundbreaking advances in fields such as computer vision, natural language processing, and speech recognition~\cite{lecun2015deep}. Over the years, the increase in scale of deep models has been instrumental in driving progress. It has now become inevitable for practitioners to embrace huge pretrained models (such as Llama~\cite{touvron2023llama} or CLIP~\cite{radford2021learning}) in the pursuit of efficient learning solutions. However, resuming training on these large models can be very demanding in terms of computational resources, and often necessitates the use of techniques that alleviate the computational and memory burden when training.
 
Low-rank training techniques~\citep{li2017learning, aghajanyan2020intrinsic, hu2021lora} can effectively finetune enormous pretrained models (up to 175 billion parameters for GPT 3.5 \citep{brown2020language}), by only tuning a few parameters (up to 10,000 times less) to later project them into the full parameter space (or only a subset of parameter space). This allows practitioners to train models using little memory. The most popular technique is Low-Rank Adaptation (LoRA) and is
extensively used to finetune large pretrained models because of its simplicity and ease of use~\cite{hu2021lora}. LoRA learns a low-rank version of the weight matrices of a given network and sums them to the existing network parameters. The adapters learned in this way can then be ``plugged'' into and ``unplugged'' from the pretrained model by simply adding or subtracting the adapter weights to or from the pretrained ones. Moreover, several adapters can be combined together \citep{chitale2023task, ilharco2022editing} by summing the adapter weights together. 
Parameter-efficient learning methods like LoRA were initially studied and successfully applied in the context of transfer learning, where the task is to optimize for a single target domain.


While pretrained models demonstrate remarkable capabilities and can be adjusted to downstream tasks in a one-step fashion, current transfer learning methods based on LoRA have not investigated the loss (catastrophic forgetting) of the capabilities of the pretrained model during the LoRA adaptation. In addition, transfer learning only considers adaptation to a single target domain, whereas in many realistic applications models should continually adapt to changing data distributions, varying domains, temporal changes, a larger set of semantic classes, etc. The field of continual learning investigates methods to address these situations, where the aim is to adapt to a changing non-IID distribution while accumulating the knowledge of all domains. However, the vast majority of continual learning literature has considered training from scratch \citep{kirkpatrick2017overcoming, li2017learning, masana2022class, de2021continual}. Recently, starting from pretrained models is gaining some attention in the community~\citep{Panos_2023_ICCV, wu2022class, wang2022learning, goswami2023fecam}. Unfortunately, most existing works do not aim to improve the performance of the pretrained model and focus on optimizing the performance on the sequence of downstream tasks. Only a few recent works consider continual improvement of the pretrained model~\citep{cossu2022continual}.

We think that with the growing importance of very large pretrained models (or foundation models) there will be a further shift towards studying parameter-efficient continual learning algorithms. Therefore, we propose to study the consequences in terms of forgetting and knowledge transfer of incremental LoRA updates. To study incremental updates of LoRA to a sequence of tasks, we propose an experimental setup in which we incrementally learn four fine-grained tasks (Cars, Flowers, Aircraft, and Birds) starting from a model pretrained on ImageNet. Contrary to many previous works, we do not aim only to achieve good performance on downstream tasks, but we also aim to maintain the capacity on the pretrained foundation task (i.e. the 1,000 ImageNet classes used for pretraining in our experiments). 


 
Our contributions are the following:
\begin{itemize}
    
    
    \item We explore the interplay between low-rank updates of pretrained model and continual learning, and we show that lower rank updates leads to less forgetting in the case of vision transformers. 
    
    \item We observe that the forgetting obtained for vision transformers (ViT) can be interpreted as a form of contextual forgetting, where the forgotten samples correspond to ImageNet classes related to the domain of the fine-grained datasets. 

    \item We evaluate incremental LoRA updates on both Vision Transformer (ViT) and Residual Network (ResNet) models and show that convolutional networks suffer far more forgetting than Vision Transformers.

\end{itemize}

\section{Related Work}

Incremental learning refers to the paradigm of updating an existing model with a continuous stream of data or tasks, while mitigating the phenomenon of catastrophic forgetting. Traditional incremental learning methods initialize the feature extractor in one of two ways: the \textit{Cold-Start} scenario, where the feature extractor is randomly initialized before the incremental learning steps \citep{kirkpatrick2017overcoming, li2017learning, magistri2024elastic, saha2020gradient, lin2021trgp}, or the \textit{Warm-Start} scenario, where the feature extractor is pre-trained on half of the available dataset before being incrementally trained on subsequent tasks~\citep{zhu2021prototype, zhu2021class,zhu2022self, petit2023fetril, goswami2023fecam}. However, the exceptional performance of large foundational models, such as Vision Transformers (ViT) \citep{dosovitskiy2021an} for vision tasks, and BERT models \citep{devlin2018bert, liu2019roberta, sanh2020distilbert} for natural language processing tasks, along with self-supervised techniques~\citep{balestriero2023cookbook} for effectively pre-training these models, has sparked an increasing interest in exploring incremental learning with \textit{Pre-Trained} architectures. Focusing on vision tasks, recent state-of-the-art methods leverage the powerful representational capabilities of ResNet architecture ~\citep{he2016deep} and ViT models, both pre-trained on the Imagenet dataset ~\citep{deng2009imagenet}, to further enhance incremental learning  outcomes. 

 \citet{Panos_2023_ICCV} adapt a pre-trained ResNet for the first task, then freeze the feature extractor and classify the classes across incremental steps using Linear Discriminant Analysis (LDA). \citet{wu2022class} propose to dynamically expand a frozen pre-trained ResNet model with new convolutional layers, fine-tuning it on new data and fuse the previous and current classifier for each new class using exemplars.   \citet{wang2022learning} propose learning to dynamically prompt a pre-trained ViT to sequentially learn tasks. \citet{goswami2023fecam} freeze a pre-trained ViT feature extractor after the first task and then use the class covariance matrix and the class means obtained by the feature extractor to perform classification on all task classes. \citet{Zhang_2023_ICCV} fine-tune the ViT feature extractor during training and, at the end of each task, calibrate the task classifiers for class-incremental evaluation using Gaussian Prototypes accumulated across all encountered tasks. 
 
 Some recent works explore the usage of low-rank training for updating a pre-trained feature extractor in incremental learning. \cite{hyder2022continual} perform \textit{Continual Learning via Low-Rank updates}, where they represent the full weight matrix as a sum of rank 1 matrices each corresponding to one task. When they learn a new task, they learn a new rank 1 matrix along with a selector vector that weights the existing matrices to better fit the current task, this method allows them to have zero forgetting in the task aware setting. \cite{wistuba2023continual} start from a pretrained model and incrementally learns per task LoRA adapters, they further select the best LoRA adapter to be applied at inference by performing task-inference via task-wise prototypes storage and retrieval. \citet{chitale2023task} propose to train a LoRA per task, then at the end of each task merge the learning low-rank parameters with the frozen feature extractor and finally tune the overall model using exemplars for class-incremental evaluation. \cite{harun2023overcoming} use LoRA in order to reduce the plasticity and find that using it has a positive impact on the stability gap.

Given the recent impressive efficacy of pre-trained models in incremental learning, current research is directed towards assessing how pretrained initialization influences the phenomenon of forgetting during the incremental learning process. These studies seek to understand the relationship between the representations learned for the pretrained (upstream) classes and those for the subsequently added (downstream) classes, as well as the extent to which pre-training methods impact overall performance. \citet{ramasesh2022effect} empirically evaluate the catastrophic forgetting on ResNet and Transformers architecture. They show that pretrained networks are less susceptible of forgetting and that their robustness improves with scale of both model and pre-training dataset size. \citet{mehta2023empirical} empirically demonstrate that pretrained initialization helps mitigate forgetting, and this effect is associated with wider minima in the loss landscape. 


\citet{lee2023pre} examine the extent to which the architecture of pretrained models and the pre-training method, whether supervised or self-supervised, influence incremental learning algorithm performance. Their analysis reveals that an under-performing incremental learning algorithm, originally designed for traditional incremental scenarios (Warm-Start or Cold-Start), can be enhanced to compete with, or even surpass, existing state-of-the-art solutions when evaluated with pretrained initialisation. Moreover, they find that combining minimal regularization with exemplars is more beneficial than stronger regularization when starting with a pretrained network. \citet{janson2022simple} shows that using a Nearest-Mean-Classifier rule on top of a frozen pretrained ViT provides competitive results with recent state-of-the-art approaches.

\citet{galashov2023continually} discuss the impact of pretrained foundation models in both supervised and self-supervised settings, demonstrating simply fine-tuning the classification heads on a top of a foundation model in task-incremental setting the final performance surpass those of models fully trained from scratch. \citet{ostapenko2022continual}  examine the efficacy of latent replay, a technique involving the replay of a network's intermediate representations across tasks. Their findings indicate that latent replay surpasses the efficiency of experience replay, which relies on replaying exemplars. The effectiveness of latent replay, however, hinges on the correlation between downstream and upstream classes. Specifically, latent replay exhibits reduced transfer effectiveness in out-of-distribution incremental scenarios, such as when transitioning from an ImageNet upstream task to a Cars downstream task. Conversely, it demonstrates notable transfer benefits in in-distribution domains, exemplified by the transition from an ImageNet upstream task to a SplitCifar100 downstream task.

\section{Low-Rank Training}

Low-Rank training for neural networks is based on the idea that while the parameter space of a network is so big, it is possible to find another parametrization of the network so that we learn weights of lower dimension and later project it back into the full parameter space. \cite{li2018measuring} found that using this technique, it is possible to train overparametrized network from scratch by sometimes only learning a few thousands of parameters that are then projected back to the full parameter dimension. \cite{aghajanyan2020intrinsic} later show that if we consider a pretrained model, the dimension of the additional parameters that is required to solve a downstream task gets lower the better the pretrain model is. \cite{hu2021lora} exploit this property and propose to learn low-rank weight matrices instead of a flat parameter vector projected back to the full parameter space. Using this technique called LoRA for Low-Rank Adaptation, they can find solutions to downstream tasks when starting from a pretrained model by learning matrices of rank as low as 1, drastically reducing the number of parameters to be learnt.

LoRA~\citep{hu2021lora} is a low-rank training method  that has been designed initially for large language models (LLMs). The principle of this method is quite straightforward, in order to ensure that the difference between the finetuned model and the pretrain model is low-rank, they parametrize this difference $\Delta W_{finetune}$ as the product of two matrices $B^{n \times r}$ and $A^{r \times m}$ with a small $r << min(n, m)$, this product has a rank less than or equal to $r$ by construction. They then scale $\Delta W_{finetune}$ by $\alpha / r$ where $\alpha$ is an hyperparameter and r the rank. While LoRA has been initially designed to work for 2 dimensional parameter tensors (matrices), which is the case for query key value parameter matrix of the transformers, it can be also applied on convolution layers of arbitrary kernel size with a small adaptation. However, since in this article we apply it on convolution layers of kernel size 1x1 (third convolution layer of each ResNet block), we can apply LoRA just the same way than on query key value matrices by viewing the $1 \times 1 \times n \times m$ kernel as an $n \times m$ matrix.  

\begin{align*}
    W & = W_{pretrain}^{n \times m} + \Delta W_{finetune}^{n \times m} \\
    & = W_{pretrain}^{n \times m} + B^{n \times r}A^{r \times m}
\end{align*}

Existing works that apply LoRA in continual learning \citep{wistuba2023continual} use one LoRA adapter per task and rehearse it at inference, \cite{chitale2023task} merge the learned LoRA and later finetune the model using stored exemplars. Approaches that store the LoRA for later rehearsal make the choice to sacrifice transfer between tasks in order to have zero forgetting. Here we want to study the forgetting induced by continually learning the pretrained model, we hence choose a natural approach which is to learn one LoRA adapter per task that we sum to the model weights at the end of each task, without storing it separately. We believe this option makes sense for the tasks we consider that are fine-grained, however we could imagine more complex schemes learning different LoRAs for different samples (for instance one LoRA per class), in order to have one LoRA of low-rank specialized to each domain, we leave this to future works.

\section{Experiments}

\subsection{Experimental Setup}

\textbf{Datasets:} We perform the continual learning experiments by learning on a sequence of 4 fine-grained classification datasets with varied topics and dataset size. Stanford Cars \citep{krause20133d} is comprised of 196 categories of cars, and contain 8,144 training images and 8,041 test images. Oxford 102 Flowers dataset \citep{nilsback2008automated} has 102 categories of flowers with around 1,000 training images. FGVC-Aircraft dataset \citep{maji13fine-grained} is comprised of a total of 10,200 images of 100 aircraft model variants. Caltech-UCSD birds dataset \citep{welinder2010caltech} contains 6033 images of 200 birds species. For all the datasets, we apply the same transformations during training and evaluation than the ones that were applied to Imagenet during training of the specific model. We consider two training settings using these datasets, one of them we coin the \textit{Short Setting} considers training on the sequence of 4 tasks Cars - Flowers - Aircraft - Birds, while the other setting we coin the \textit{Long Setting} considers training on a two repetitions of this sequence, and splits each dataset in two random parts, to create a sequence of 8 training experiences by revisiting the previous tasks.

\textbf{Models:} To get the pretrained models, we use the timm \citep{rw2019timm} library and fetch pretrained model from the HuggingFace database. In particular, and in order to provide a more complete analysis, we use two models, one ViT \citep{dosovitskiy2020image} with patch size 16 (HuggingFace model id: "timm/vit\_base\_patch16\_224.augreg\_in1k") , with a total of 86M parameters, and one ResNet-50 \citep{he2016deep} (HuggingFace model id: "timm/resnet50.a1\_in1k"), with 21M parameters, both pretrained on Imagenet 1k \citep{deng2009imagenet}. Both model have similar performance on the Imagenet test set, with the ViT having $79.14\%$ accuracy and the ResNet-50 $80.04\%$ Top-1 accuracy.

\textbf{Low-Rank Training:} In order to perform low-rank incremental training, we setup a new LoRA adapter for each task that we merge into the model weights at the end of training each task by simply summing the adapter weights to the model weights. For the ViT, we choose to apply the adapter on every attention matrix (query, key, value weight matrix), which is a classical process. For the ResNet, it is less obvious how to choose the parameters to adapt since adapting every convolutional layer would results in prohibitive amount of adapted parameters. In order to provide a fair comparison with the ViT, we choose to adapt only the last convolutional layer of each block ($1 \times 1$ convolution), which results in an amount of modifiable parameters that roughly matches the one of the ViT attention matrices (1/4th of the total number of parameters of the network). We then vary the rank of the LoRA which in turns varies the number of trainable parameters, we provide in the appendix a comparison on how this number of trainable parameters evolve linearly but with different slopes for the chosen ResNet and ViT. We insist here that although the number of trainable parameter varies with the rank, the number of modifiable parameters does not depend on the rank and is always fixed once we have chosen the type of layer to modify. Additionally, we tune the parameter alpha of the LoRA to be equal to two times the LoRA rank, we make this choice in order to be able to work at fixed learning rate while fairly comparing results across ranks, since fixing alpha and increasing the rank leads to smaller learning rate for bigger ranks, which also has an effect on learning and forgetting.

\textbf{Optimization:} We use the AdamW optimizer \citep{kingma2014adam} with learning rate of $1e-3$, a batch size of 32 and a cosine learning rate scheduler with warmup that we restart for each task, we choose the warmup steps to be always equal to 1/18 th of the total number of finetuning steps for one task, and we roughly tune the number of epochs for each task so that we train for a bit more epochs on tasks that are very small (in order to get a decent number of training iterations and converging training accuracy).

\textbf{Reproductibility:} To perform our continual learning experiments, we use the avalanche library \citep{lomonaco2021avalanche, carta2023avalanche}, which is an open source library available at \url{https://github.com/ContinualAI/avalanche}. The code to reproduce our experiments is available at \url{https://github.com/AlbinSou/lora_cl_analysis}.

\subsection{Results}

\textbf{Impact of forgetting the pretrain task:} To assess the impact of forgetting of the pretrain task on further transfer capacity, we compare the results of finetuning the pretrained model directly on each fine-grained task (i.e. \textit{Direct Transfer}) versus finetuning it continually, by taking the checkpoint at the end of learning each task to start training on the subsequent task. In order to fairly compare it to later experiments done with LoRA, we finetune only the query key value matrices for ViT and the third convolutional layer of each block for ResNets. We present the results of this experiment in Table~\ref{tab:onestep}. We see that after learning 'Cars' and 'Flowers' the network only obtains a 30\% performance on 'AirCraft', whereas a direct adaptation to 'Aircarft' would yield 58\%, showing that much of the pretrained knowledge useful for airplane classification has been lost by the two finetuning stages. In order to avoid that while still continually learning the pretrained model, we need to avoid forgetting of the pretrain task. 

\begin{table}[h!]
    \centering
    \begin{tabular}{ccc}
         \hline
         \textbf{Task} & \textbf{Continual FT} & \textbf{Direct Transfer}  \\
         \hline
         Cars & 83.0 & 83.0 \\
         Flowers & 86.1 (-\textbf{4.2}) & 90.3 \\
         AirCraft & 30.0 (-\textbf{28.6}) &  58.6 \\
         Birds & 36.8 (-\textbf{36.9}) & 73.7 \\
         \hline
    \end{tabular}
    \caption{Comparison between the (first) accuracy obtained on downstream tasks when performing continual finetuning on the task sequence Cars - Flowers - AirCraft - Birds versus when performing direct transfer learning on each dataset, starting from the pretrained ViT model.}
    \label{tab:onestep}
\end{table}

\textbf{Does the LoRA rank impact forgetting?} We experiment with various ranks of the LoRA adapter that we train and merge for each task. We then report the accuracy and forgetting on the pretrained task (Imagenet 1k \citep{deng2009imagenet}) as well as on the downstream tasks. Additionally, we apply LwF \citep{li2017learning}, a continual learning method that could realistically be applied without access to the pretraining task data, on top of the low-rank learning procedure, in order to confirm that the gains obtained by low-rank learning are orthogonal to the gains obtained using existing continual learning method. We experiment with ranks in the range [1, 2, 4, 6, 8, 16, 32] for ViT and [1, 2, 3, 4, 5, 6, 7] for ResNet (See Discussion in Appendix Section \ref{sec:params}). 

We present the results for both continual low-rank finetuning and LwF in Figure~\ref{fig:lora_ft} and Figure~\ref{fig:lora_lwf} respectively, and the detailed results for finetuning in Table~\ref{tab:table_forgetting}, where we also include one baseline \textbf{Head} that freezes the backbone and tunes the head only, and one baseline \textbf{Full} that fully finetunes the subset of parameters that we apply LoRA on. In Figure~\ref{fig:lora_ft} we see that the rank chosen for the adapter has a significant impact on the forgetting of both the pretraining task and on the downstream task Cars (this conclusion generally holds for all tasks in the sequence), this impact also seems more important for ViT than for ResNet. We see that while learning an adapter of higher rank, the accuracy on the downstream task is positively affected (at least up to the maximum rank considered here), but choosing a very low-rank already reaches very satisfying accuracy compared to training only the output head, providing a good stability-plasticity tradeoff. In Figure \ref{fig:lora_lwf}, we see that when applying LwF in combination with LoRA, we still see a big discrepancy between LoRA adapters of different ranks, although the forgetting is still drastically reduced for all of the ranks. These results indicate that the rank has an impact on forgetting that is orthogonal to the one of LwF. Interestingly, by combining ViT with an adapter of rank 1 and LwF, we even witness some slight backward transfer on Imagenet when learning on Cars with the accuracy increasing by a small amount of 0.1\%. We also see in Table ~\ref{tab:table_forgetting} that the gains in plasticity obtained by training adapter of higher rank do not manage to convert in higher overall average accuracy since it is counterbalanced by too high forgetting, it results in a final average accuracy that diminishes with the rank.

\tableforgettingshortnew{Final average accuracy and forgetting after learning on the \textit{Short Setting}, depending on the rank, for ViT (Left) and ResNet (Right). Results are averaged over 4 runs using different seeds.}

\figloraftshort{10 0 70 50}{0.95}{Test Accuracy on the pretrain task Imagenet1k (Left) and on the downstream task Cars (Right) when fine tuning on the \textit{Short Setting} with multiple LoRA ranks, using ViT network (Blue) and ResNet-50 (Green). Results are averaged over 4 runs using different random seed.}

\figloralwf{10 0 70 50}{0.95}{Test Accuracy on the pretrain task Imagenet1k (Left) and on the downstream task Cars (Right) when training on the \textit{Short Setting} with multiple LoRA ranks in combination with the LwF method, using ViT network (Blue) and ResNet-50 (Green). Results are averaged over 4 runs using different random seed.}

\textbf{Interpretation of the forgetting of the pretraining task:} Since the tasks that we choose have a very specific domain and the pretrain task covers a wide variety of domains with some categories that lie in the domain of the fine-grained tasks, we try to interpret the forgetting of the pretraining task by looking at the most forgotten categories of Imagenet1k. Whenever learning on a new downstream task, we report the new errors that the model make on Imagenet. In this experiment, we exclusively consider \textit{direct transfer} from the pretrained model using a LoRA adapter, to maintain clarity and avoid confounding variables. We present these differences in error vector by using two methods, the first method simply considers the histogram of the most affected categories, while the second one considers comparing the distributions of the semantic distances between the category name and a target word that describes the domain of the fine-grained dataset. We chose to report the Wu-Palmer semantic similarity which is widely used in the field of Natural Language Processing~\citep{wu1994verb}. We present the distribution of this similarity score for the categories of new errors samples and compare it to the distribution of similarities of all Imagenet1k categories, which can be seen as the distribution of random errors.

In Figure \ref{fig:forget_hist}, we show the most forgotten Imagenet categories after learning on the Cars dataset, using ViT with a LoRA adapter of rank 32, with a total forgetting of 7\% on Imagenet test set. Looking at the category names, we see that the most forgotten one is "sports cars" from Imagenet, which indeed is very related to the Cars dataset domain. While not all categories in the histogram are related with cars, there is a good amount of them that are, among them "grille", "tow trucks", "convertible", "minivan", "limousine", "pickup" are all present in the top 20 forgotten categories, which provides strong evidence that there is contextual forgetting. The same conclusion holds for Aircrafts, where the first forgotten Imagenet category is "airliner". In Figure~\ref{fig:forget_kde}, we see that while Imagenet categories are generally lying in two "modes" of similarity with the word plane and cars, the newly forgotten categories mainly lie in the mode that is most similar to the target word, with a small spike on words that are very similar, which corresponds to the most forgotten categories that we observe in Figure~\ref{fig:forget_hist}. Interestingly, we do not observe such contextual forgetting for ResNet-50 model (See Appendix Figure~\ref{fig:context_resnet}). We hypothesise that this contextual forgetting happens because of a bigger feature drift for images which category is related to the downstream task at hand, which results in bigger forgetting due to the missing head adaptation, although the network might now be better at distinguishing these categories since it "improved" this part of the feature space.

\figbarplot{10 10 10 10}{0.90}{20 most forgotten categories of Imagenet1k after learning on Stanford Cars (Left) and Aircraft (Right), using direct transfer from the pretrained ViT model with a LoRA adapter of rank 32}

\figkde{20 10 20 20}{0.9}{Wu-Palmer similarity between the forgotten category names an the word "Car" after learning on Cars dataset (Left) and with the word "Plane" after learning on Aircraft dataset (Right). Comparison of the similarity distribution of all Imagenet categories versus the forgotten categories (weighted by the amount of test images forgotten in this category), when using direct transfer from the pretrained ViT model with a LoRA adapter of rank 32}

Although we observed contextual forgetting quite clearly for the ViT model, we did not find evidence of it in the ResNet-50 model. In Figure~\ref{fig:context_resnet}, we perform the same analysis as we did in Figure~\ref{fig:forget_hist} and \ref{fig:forget_kde} but for ResNet-50 instead of ViT. In order to have the most chances to observe this contextual forgetting, we choose the finetuning run that had the lowest forgetting using a LoRA of rank 1, which results in 60\% accuracy on Imagenet after learning on Cars. However, even when doing so, we find no evidence of contextual forgetting for this network. Indeed, for that network, the most forgotten categories do not seem very related to the downstream task that has been trained, and the distribution of new errors almost matches the distribution of all Imagenet categories which shows no bias towards forgetting of classes semantically related to the downstream task domain.

\figkderesnet{20 0 0 0}{0.95}{(Left) 20 most forgotten classes of Imagenet1k after learning on Stanford Cars, (Right) Wu-Palmer similarity between the forgotten category names an the word "Car". Comparison of the similarity distribution of all Imagenet categories versus the forgotten categories (weighted by the amount of test images forgotten in this category), when using direct transfer from the pretrained ResNet-50 model with a LoRA adapter of rank 1}

\textbf{Can we improve the forward transfer capabilities of the base model?:} In order to answer this question, we use the \textit{Long Setting} described in the Dataset section, which revisits each fine-grained task one more time, but can only learn with half of the data every time. In order to see if forward transfer is possible we are interested in looking at the accuracy reached the second time the task is encountered and compare it to the accuracy reached when the task is encountered the first time. If we can reach a better accuracy when encountering the second part of the dataset, it means that forward transfer is effective.

In Figure~\ref{fig:lora_ft_long}, we show the results on the \textit{Long Setting}. Curiously in that case, we get more occurrences of backward transfer on Imagenet1k, where the accuracy initially goes down but goes back up at some point in the stream. We also notice that there is knowledge transfer between both half of the downstream task dataset both for ViTs and ResNet. For instance for the Aircraft task, the first accuracy sets up at around 45\% for ViT and 30\% for ResNet, and jumps to around 52\% for ViT and 45\% for ResNet when the task is encountered for the second time. This hints at the fact that we actually managed to improve the forward transfer capabilities of the model by training on the first half of the dataset so that we can reach better accuracy on the second half, which is an encouraging step regarding the possibility to continually improve these models. We also see in Figure~\ref{fig:lora_avg} that the final average accuracy over all tasks is not affected by the \textit{Long Setting} for ViTs trained with very low-rank adapters, and even improved for ViTs trained with adapters of higher rank. In comparison, the accuracy for ResNet is overall negatively affected (around 3\% drop in average accuracy between the two settings).

In Figure~\ref{fig:lora_ft_long_8020}, we show the accuracy on the downstream task Cars for two modified versions of the \textit{Long Setting}, one where we learn first 20\% of the data of each task and later 80\%, and another where we learn first 80\% and later 20\%. We see that when learning on 20\% in the second round, the model is able to reach accuracies much higher (70\%) than when learning on 20\% in the first round (40\%), which shows that important forward transfer occurred. Surprisingly, both reach similarly high final accuracy which is only slightly lower than the accuracy obtained in the 50-50 setting (see Appendix Figure~\ref{fig:lora_additional_long}).

\figloraftlong{10 0 70 50}{0.95}{Test Accuracy on the pretrain task Imagenet1k (Left) and on the downstream task Aircraft (Right) when fine tuning on the \textit{Long Setting} with multiple LoRA ranks, using ViT network (Blue) and ResNet-50 (Green). Results are averaged over 4 runs using different seeds}

\figloraftavg{10 0 70 50}{0.95}{Average test accuracy across all seen tasks on the \textit{Short Setting} (Left) and on the \textit{Long Setting} (Right) when fine tuning with multiple LoRA ranks, using ViT network (Blue) and ResNet-50 (Green). Results are averaged over 4 runs using different seeds}

\figloraftlongtwenty{10 0 70 50}{0.95}{Test Accuracy on the downstream task Cars on two modified versions of the \textit{Long Setting} with 20\%-80\% of the data (Left) and 80\%-20\% of the data present in the first and second occurences of the downstream tasks respectively,for multiple LoRA ranks, using ViT network (Blue) and ResNet-50 (Green).}


\textbf{Representation strength in the class incremental learning setting:} While the previous results are provided in the task-incremental learning (task-IL) setting. We propose here to evaluate the representation obtained after training and merging the LoRAs under the class-incremental (class-IL) setting, in order to determine whether the same conclusions apply. To do so, we perform linear probing \cite{alain2017understanding, davari2022probing}, using the full training data, on the representations that were learned in the task-IL experiments, but this time report the task agnostic (class-IL) accuracy obtained for multiple ranks when using this linear probe, as well as the accuracy on Imagenet test set and the forgetting on this same dataset. Note that this is done purely as an evaluation purpose to estimate the representation strength and the forgetting suffered by the representation, this method cannot be applied without access to the previous data. We present the results in Table~\ref{tab:linear_probe}. We see that the accuracies obtained are surprisingly high and sometimes surpass the ones obtained when performing task aware inference. This is due to the combination of the learning of a linear probe which is able to adapt the head to the drifting representation, and to the fine-grained nature of the tasks which makes the task-inference problem quite easy.

\begin{table}[h!]
    \centering
    \begin{tabular}{cccc}
         \hline
         \textbf{Rank} & \textbf{Average Accuracy} & \textbf{Imagenet Accuracy} & \textbf{Imagenet Forgetting} \\
         \hline
         1 & 70.3 & 70.9 & 8.2 \\
         2 & 70.1 & 67.8 & 11.3 \\
         4 & 69.3 & 66.5 & 12.6 \\
         6 & 68.9 & 64.6 &  14.5 \\
         8 & 67.8 & 62.8 & 16.3 \\
         16 & 65.6 & 58.2 & 20.9 \\
         32 & 61.1 & 49.8 & 29.3
    \end{tabular}
    \caption{Average task-agnostic accuracy (averaged over all tasks) of the linear probe on the representations of the final checkpoints from the \textit{Short Setting} experiments for ViT.}
    \label{tab:linear_probe}
\end{table}

\section{Conclusion}

In this article we explored the possibility of continually training large pretrained models by finetuning them on small, fine-grained downstream tasks using low-rank adapters that we merge at the end of each task. While doing so, we monitor the performance of the pretraining task in order to determine how much the learning of downstream tasks affects it. We make a few of interesting observations.

Firstly, we see that varying the rank of the low-rank adapter has an important impact on forgetting of both the pretrain task and the downstream tasks, and that in general forgetting diminishes when the rank of the adapter is low. These gains in stability are still present when combined with the existing continual learning method LwF, which is a method that can be applied without access to the pretrain task dataset.

Secondly, we observe that when finetuned on fine-grained downstream tasks with a specific domain, ViTs exhibit \textit{contextual forgetting} in which the pretrain task categories that are forgotten are semantically related to the downstream tasks on which it has been finetuned. This observation could help design new methods to tackle forgetting, for instance by choosing to replay classes that are semantically related to the downstream task (i.e replaying imagenet "sports cars" and "truck" categories more frequently than other categories when learning on the Cars dataset).

Lastly, we observed many differences in the application of LoRA to ViTs and ResNets. In particular, we do not observe \textit{contextual forgetting} in ResNets, but we do observe a drop in performance for ResNet training in the \textit{Long Setting} whereas ViT conserves similar accuracies to the \textit{Short Setting}. In general, we also find the forgetting in ResNet to be less affected by the rank of the LoRA adapter, and to be higher in general than the one observed in ViTs when using LoRA adapters of low rank, while they have lower forgetting than the ViT when performing full-finetuning. This suggests that transformer-like architectures are more adept at accumulating knowledge through incremental low-rank updates.

Overall, we believe that using per-task low-rank updates in order to incrementally improve pretrained vision transformers models is a promising direction and we hope that future research can exploit our analysis to further explore it.

\minisection{Acknowledgements}
We acknowledge projects TED2021-132513B-I00 and PID2022-143257NB-I00 funded by MCIN/AEI/10.13039/501100011033, by European Union NextGenerationEU/PRTR, by ERDF A Way of Making Europe, and by Generalitat de Catalunya CERCA Program. This work was supported by funding by the European Commission Horizon 2020 grant \#951911
(AI4Media). 

\newpage

\bibliography{collas2024_conference}
\bibliographystyle{collas2024_conference}

\appendix
\section{Appendix}

\subsection{Comparison between the number of affected parameters in ResNet and ViT models}

\label{sec:params}

\begin{figure}
    \centering
    \includegraphics[width=0.6\textwidth]{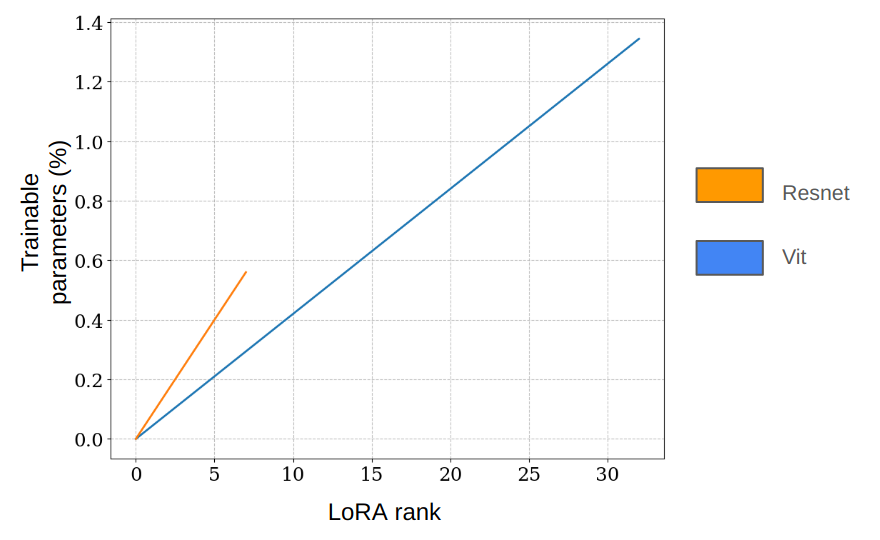}
    \caption{Comparison of the number of trainable parameters as a function of the LoRA rank for both ResNet-50 and ViT (in \% of the total parameters).}
    \label{fig:params}
\end{figure}

In order to fairly compare the use of different LoRA ranks between ViTs and ResNets, we chose to allow the LoRA adapters on ResNet to modify and to train roughly the same amount of parameters than for ViT (in terms of \% of total number of paramters). This results in 21M modifiable parameters for ViT (out of 86M total) and 5M modifiable parameters for ResNet-50 (out of 21M total), so for both networks it corresponds to approximately 1/4 of the total number of parameters. While the modifiable parameters do not depend on the rank (even with rank 1 every parameter of the chosen layers can be modified), the number of trainable parameters (which is an underparametrization of the number of modifiable parameters), does depend on the rank. We provide a comparison of the number of trainable parameters as a function of the rank for both ResNet-50 and ViT in Figure~\ref{fig:params}. We choose a range of rank of [1, 2, 3, 4, 5, 6, 7] for ResNet-50 and [1, 2, 4, 6, 8, 16, 32] for ViT. We decided to expand the range more for ViT based on the observation that forgetting was lower for lower ranks of LoRA in combination with ViT, whereas it was quite high forgetting for ResNet-50 even with low-rank, and increasing the rank makes it even worse as seen in Figure~\ref{fig:lora_ft}.

\subsection{Additional results}

\subsubsection{Other Tasks}

We include in this section more plots with the accuracies of all the considered tasks.

\begin{figure}[H]
  \centering
  \begin{subfigure}[b]{0.4\textwidth}
    \includegraphics[trim=10 0 70 50, clip, width=0.95\linewidth]{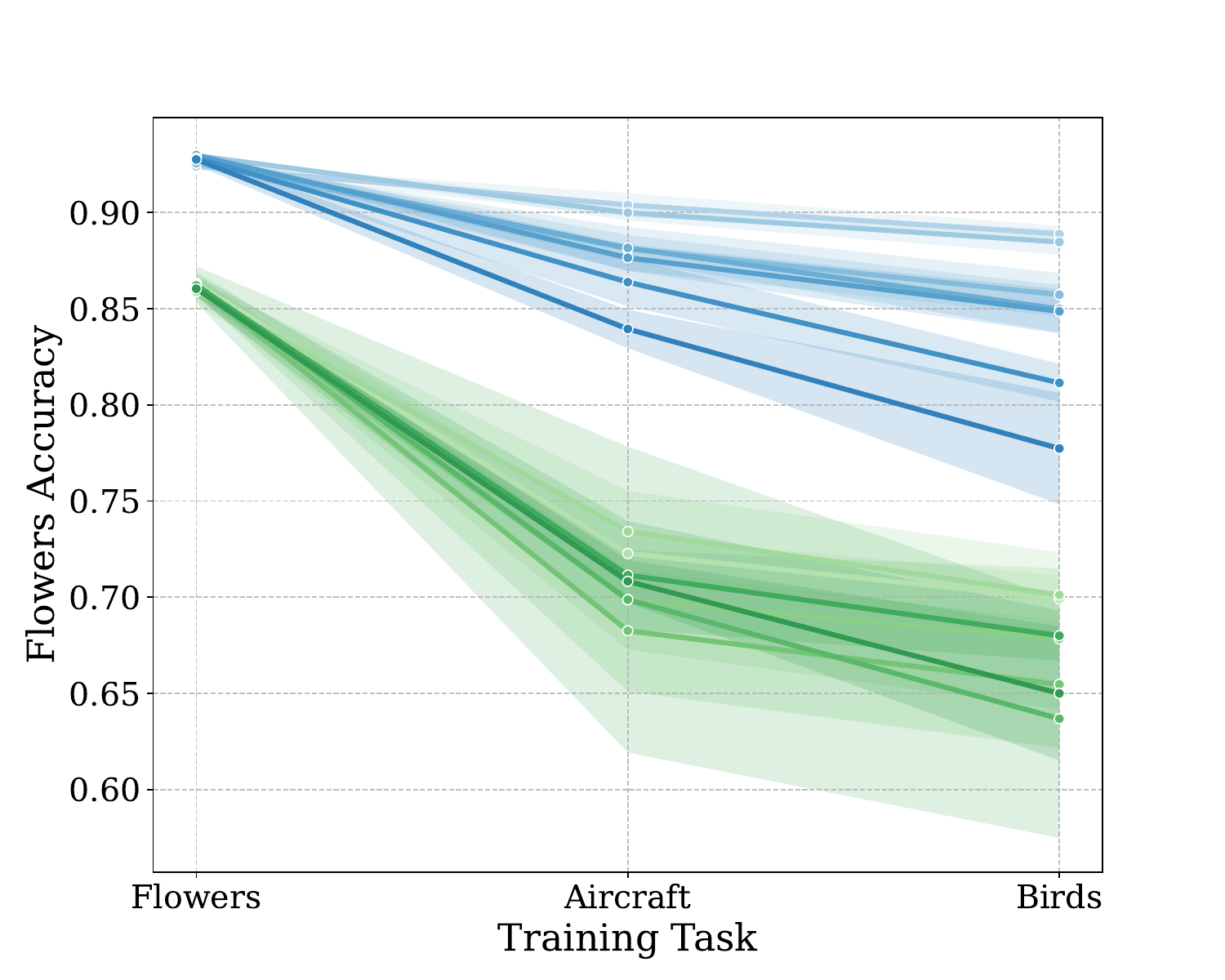}
  \end{subfigure}%
  \begin{subfigure}[b]{0.4\textwidth}
    \includegraphics[trim=10 0 70 50, clip, width=0.95\linewidth]{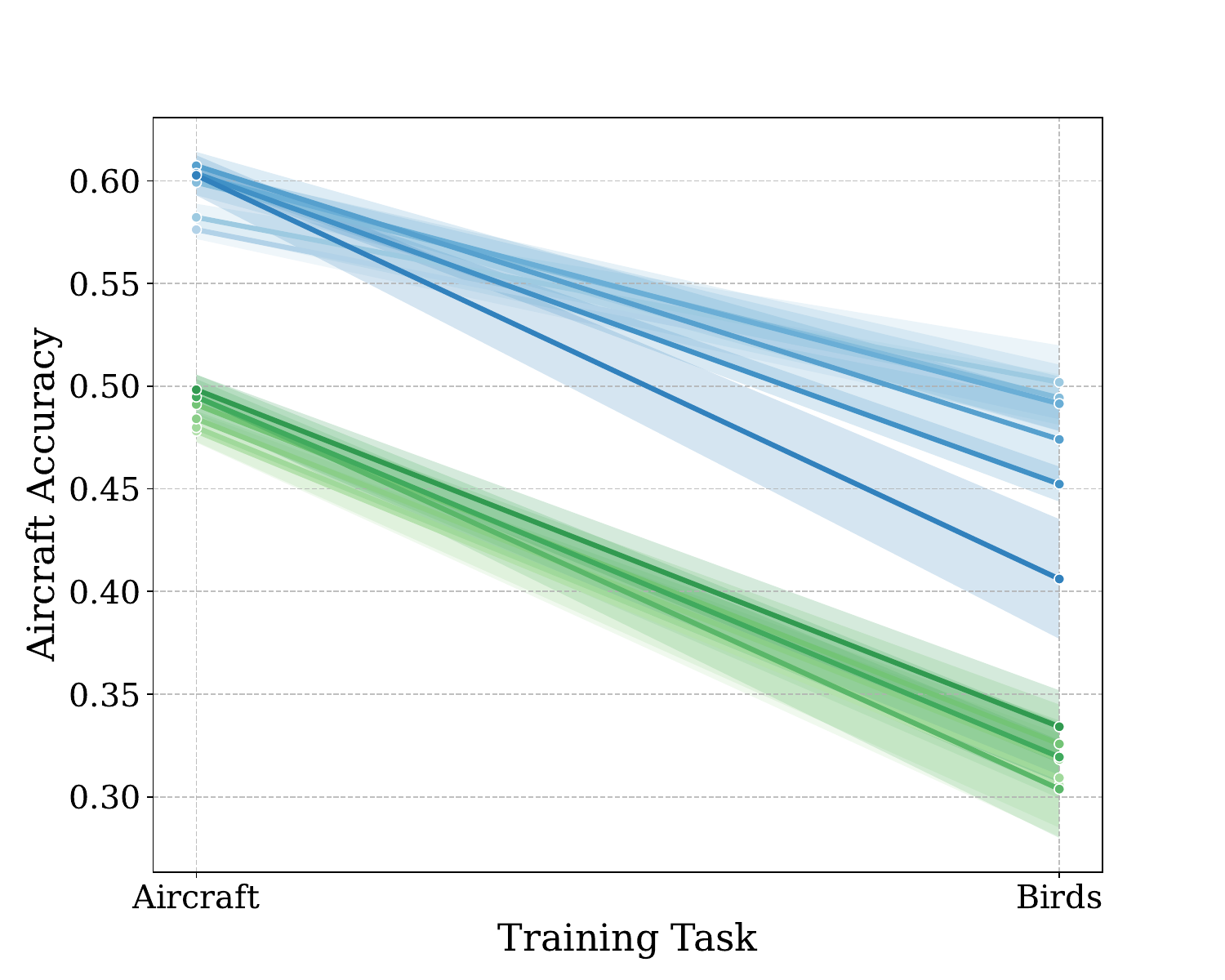}
  \end{subfigure}%
  \begin{subfigure}[b]{0.2\textwidth}
    \includegraphics[trim=400 0 0 0, clip, width=0.85\linewidth]{figures/legend_nofull.pdf}
  \end{subfigure}
  \caption{Test Accuracy on the Flowers dataset (Left) and on the Aircraft dataset (Right) when training on the \textit{Short Setting} with multiple LoRA ranks, using ViT network (Blue) and ResNet-50 (Green). Results are averaged over 4 runs using different random seed}
  \label{fig:lora_additional_short}
\end{figure}

\begin{figure}[H]
  \centering
  \begin{subfigure}[b]{0.4\textwidth}
    \includegraphics[trim=10 0 70 50, clip, width=0.95\linewidth]{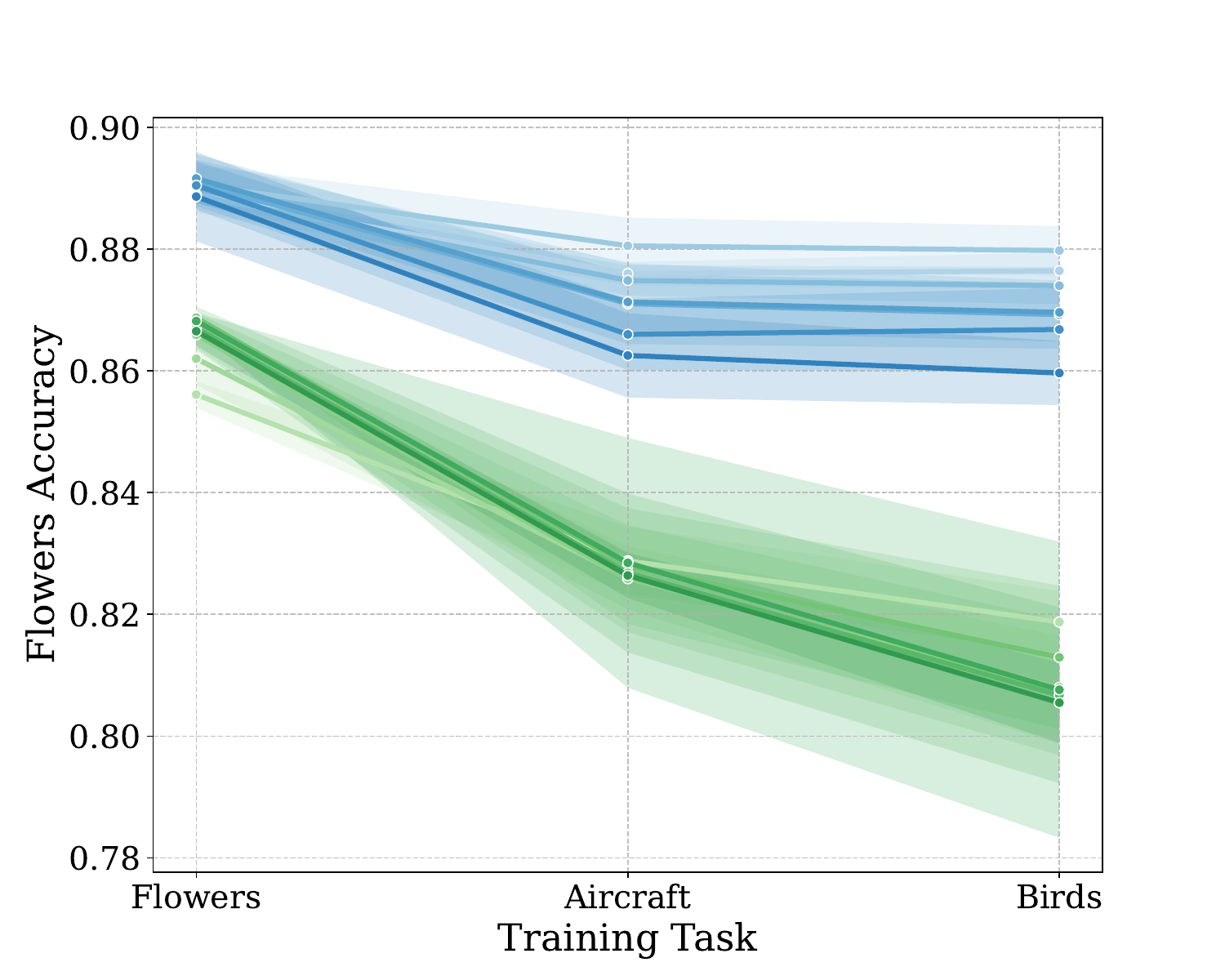}
  \end{subfigure}%
  \begin{subfigure}[b]{0.4\textwidth}
    \includegraphics[trim=10 0 70 50, clip, width=0.95\linewidth]{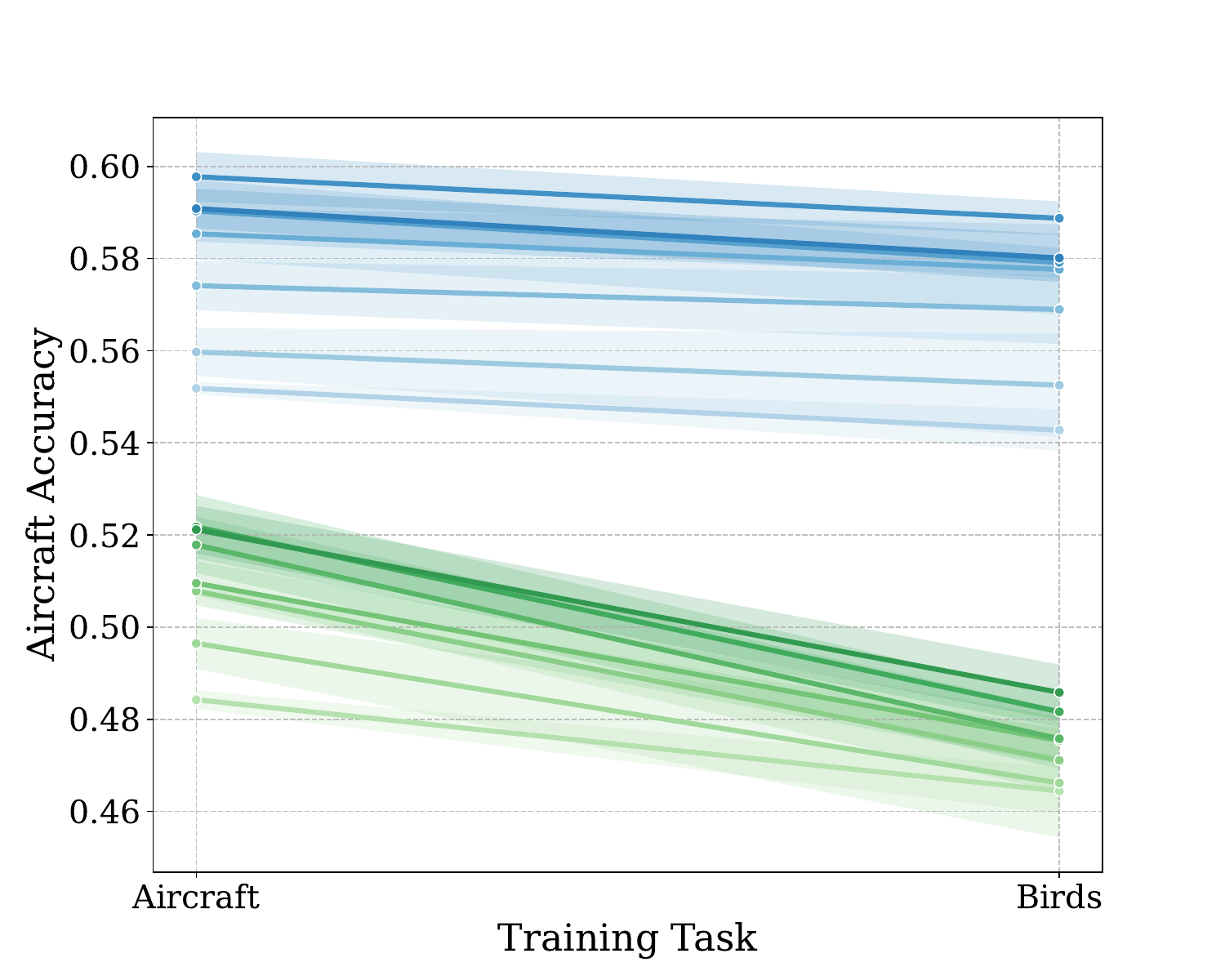}
  \end{subfigure}%
  \begin{subfigure}[b]{0.2\textwidth}
    \includegraphics[trim=400 0 0 0, clip, width=0.85\linewidth]{figures/legend_nofull.pdf}
  \end{subfigure}
  \caption{Test Accuracy on the Flowers dataset (Left) and on the Aircraft dataset (Right) when training on the \textit{Short Setting} with multiple LoRA ranks in combination with the LwF method, using ViT network (Blue) and ResNet-50 (Green). Results are averaged over 4 runs using different random seed.}
  \label{fig:lora_additional_lwf_short}
\end{figure}

\begin{figure}[H]
  \centering
  \begin{subfigure}[b]{0.4\textwidth}
    \includegraphics[trim=10 0 70 50, clip, width=0.95\linewidth]{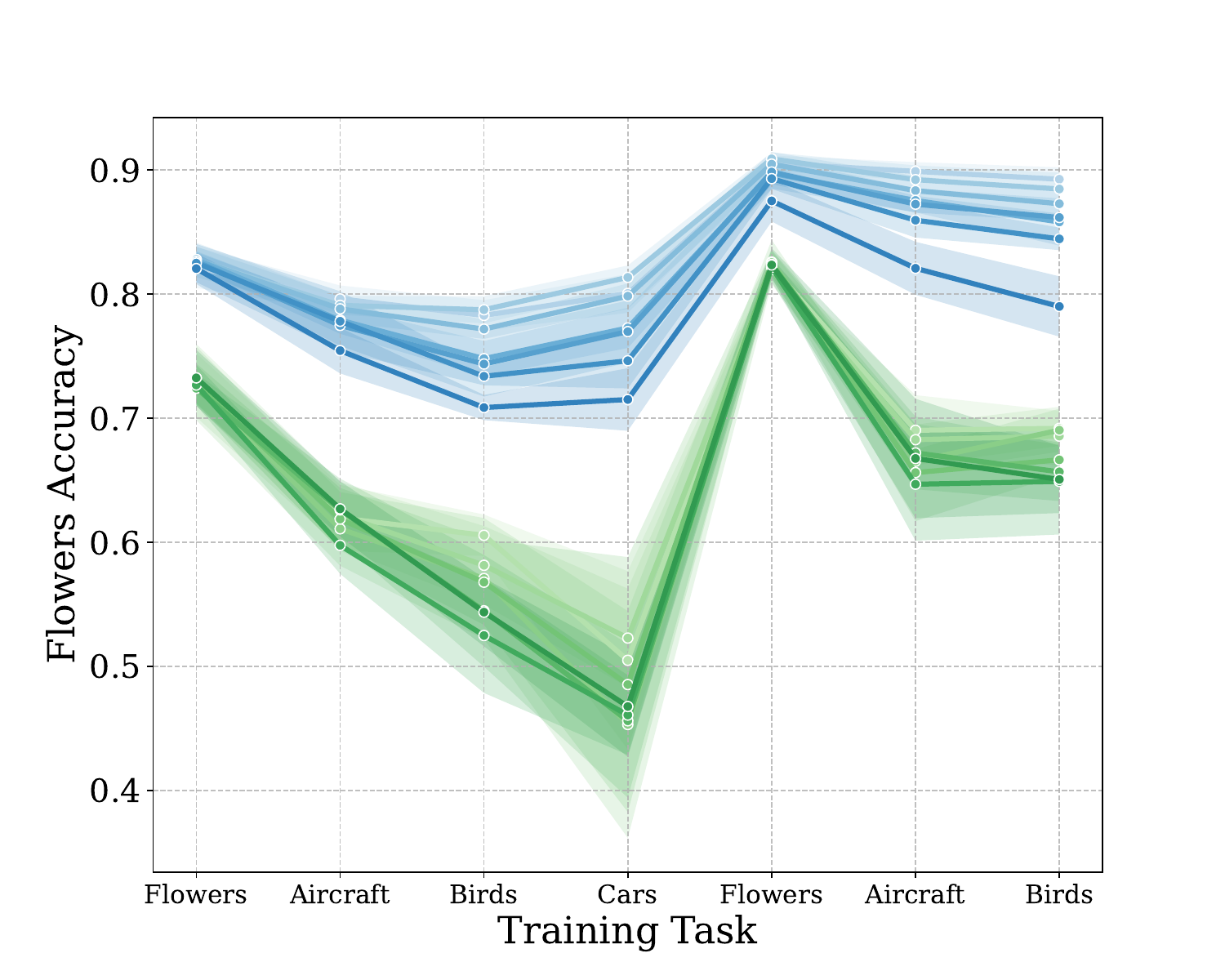}
  \end{subfigure}%
  \begin{subfigure}[b]{0.4\textwidth}
    \includegraphics[trim=10 0 70 50, clip, width=0.95\linewidth]{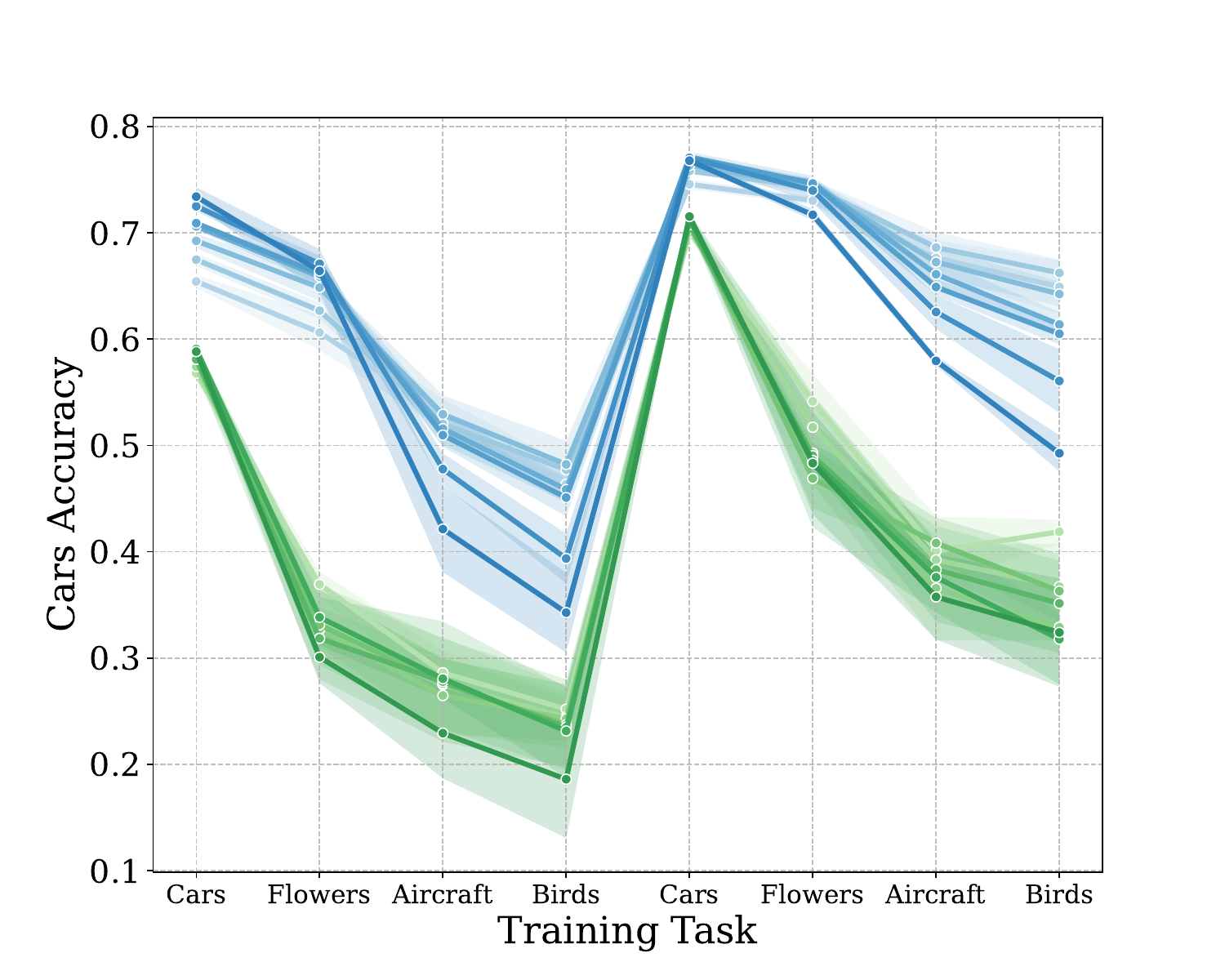}
  \end{subfigure}%
  \begin{subfigure}[b]{0.2\textwidth}
    \includegraphics[trim=400 0 0 0, clip, width=0.85\linewidth]{figures/legend_nofull.pdf}
  \end{subfigure}
  \caption{Test Accuracy on the Flowers dataset (Left) and on the Cars dataset (Right) when training on the \textit{Long Setting} with multiple LoRA ranks, using ViT network (Blue) and ResNet-50 (Green). Results are averaged over 4 runs using different random seed.}
  \label{fig:lora_additional_long}
\end{figure}

\begin{figure}[H]
  \centering
  \begin{subfigure}[b]{0.4\textwidth}
    \includegraphics[trim=10 0 70 50, clip, width=0.95\linewidth]{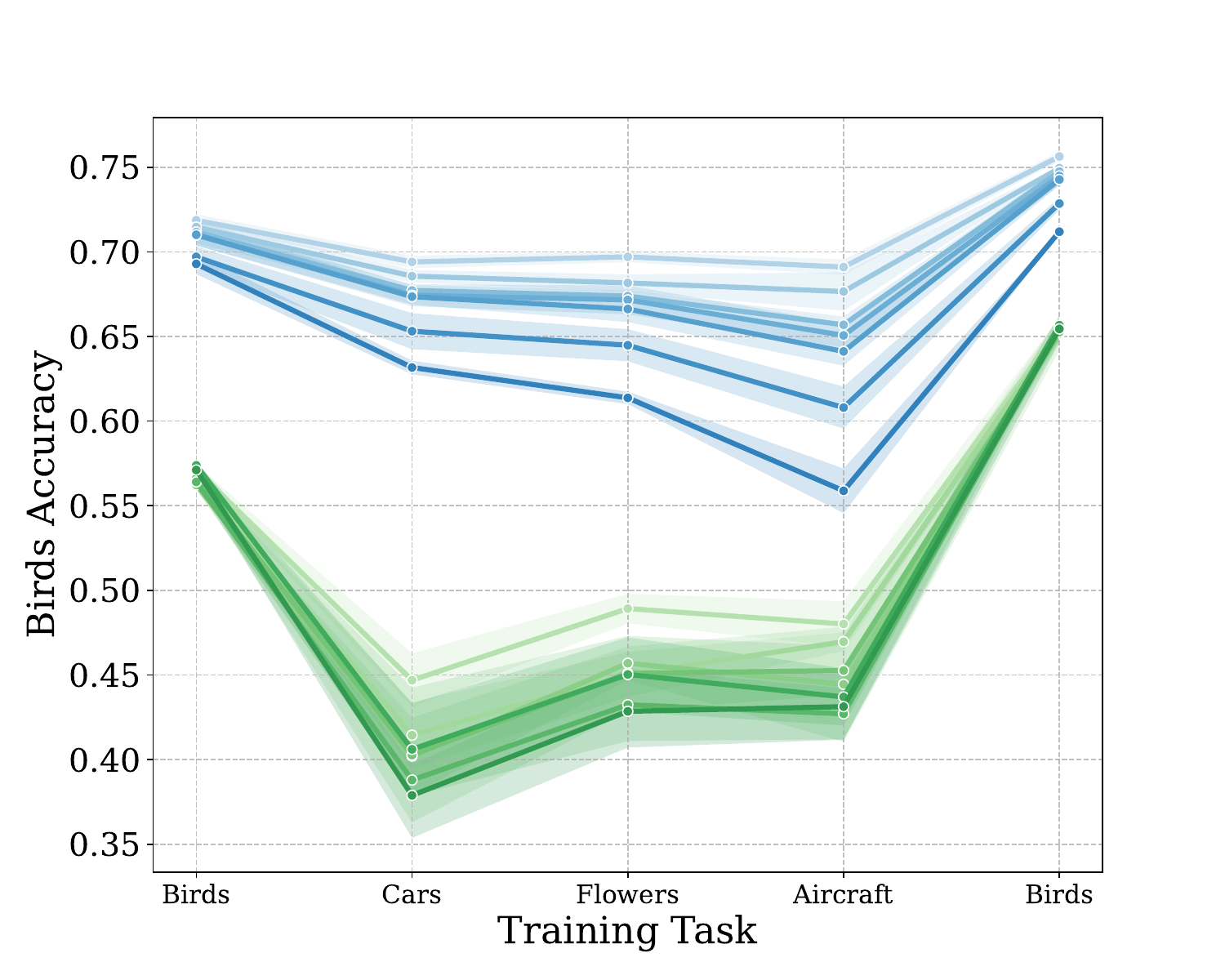}
  \end{subfigure}%
  \caption{Test Accuracy on the Birds dataset when training on the \textit{Long Setting} with multiple LoRA ranks, using ViT network (Blue) and ResNet-50 (Green). Results are averaged over 4 runs using different random seed.}
  \label{fig:lora_additional_birds_long}
\end{figure}

\subsubsection{Other Task Order}

In this section, we present results for another task order. Here we consider the order Imagenet - Aircraft - Birds - Flowers - Cars. We chose to not mix the task orders so that we can perform meaningful interpretation of the task forgetting which would be lost information if we average over all orders. For instance, we observe in general higher forgetting on the Cars dataset when we train on Aircraft dataset and reversely more forgetting on Aircraft dataset when training on Cars dataset.

\begin{figure}[H]
  \centering
  \begin{subfigure}[b]{0.4\textwidth}
    \includegraphics[trim=10 0 70 50, clip, width=0.95\linewidth]{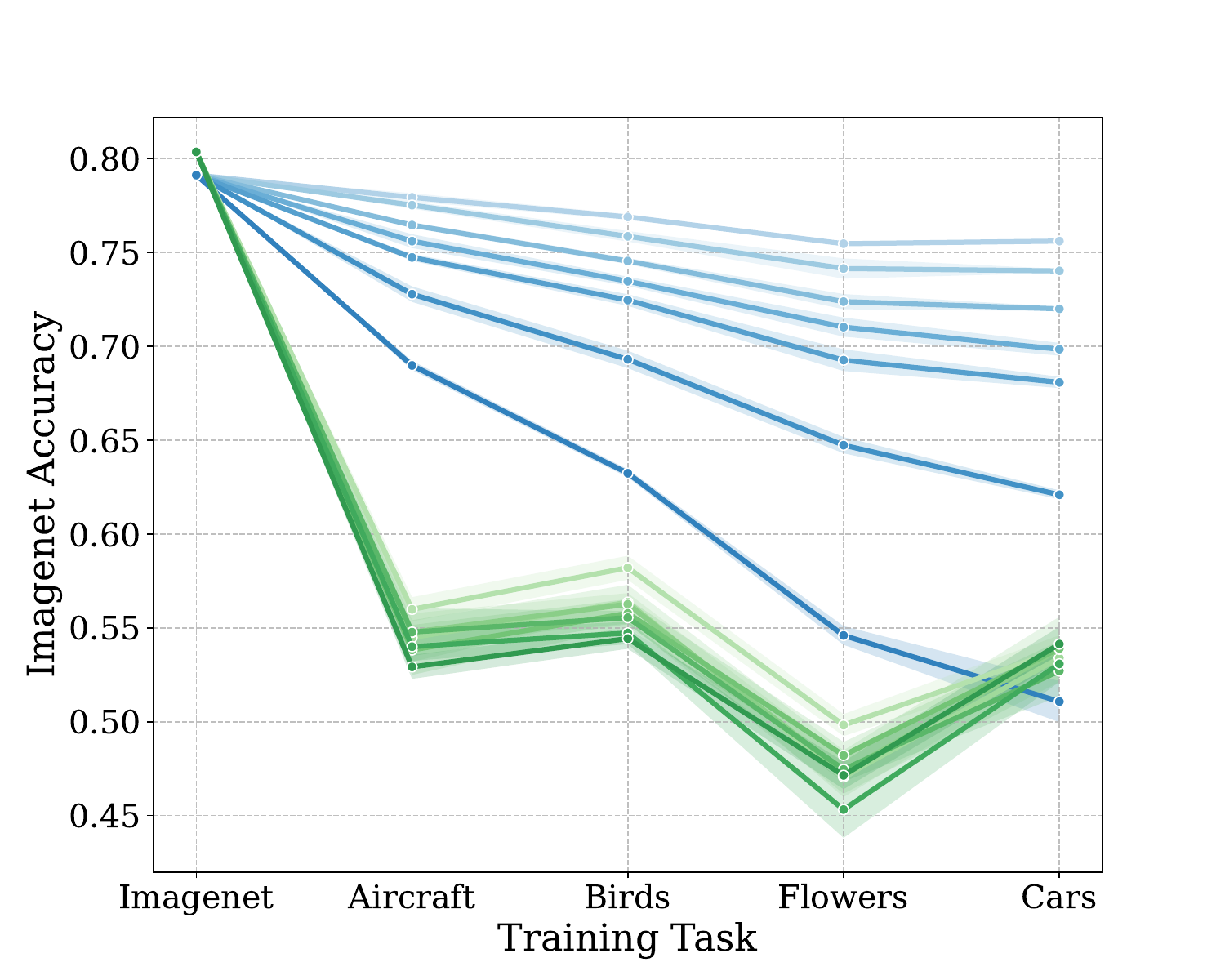}
  \end{subfigure}%
  \begin{subfigure}[b]{0.4\textwidth}
    \includegraphics[trim=10 0 70 50, clip, width=0.95\linewidth]{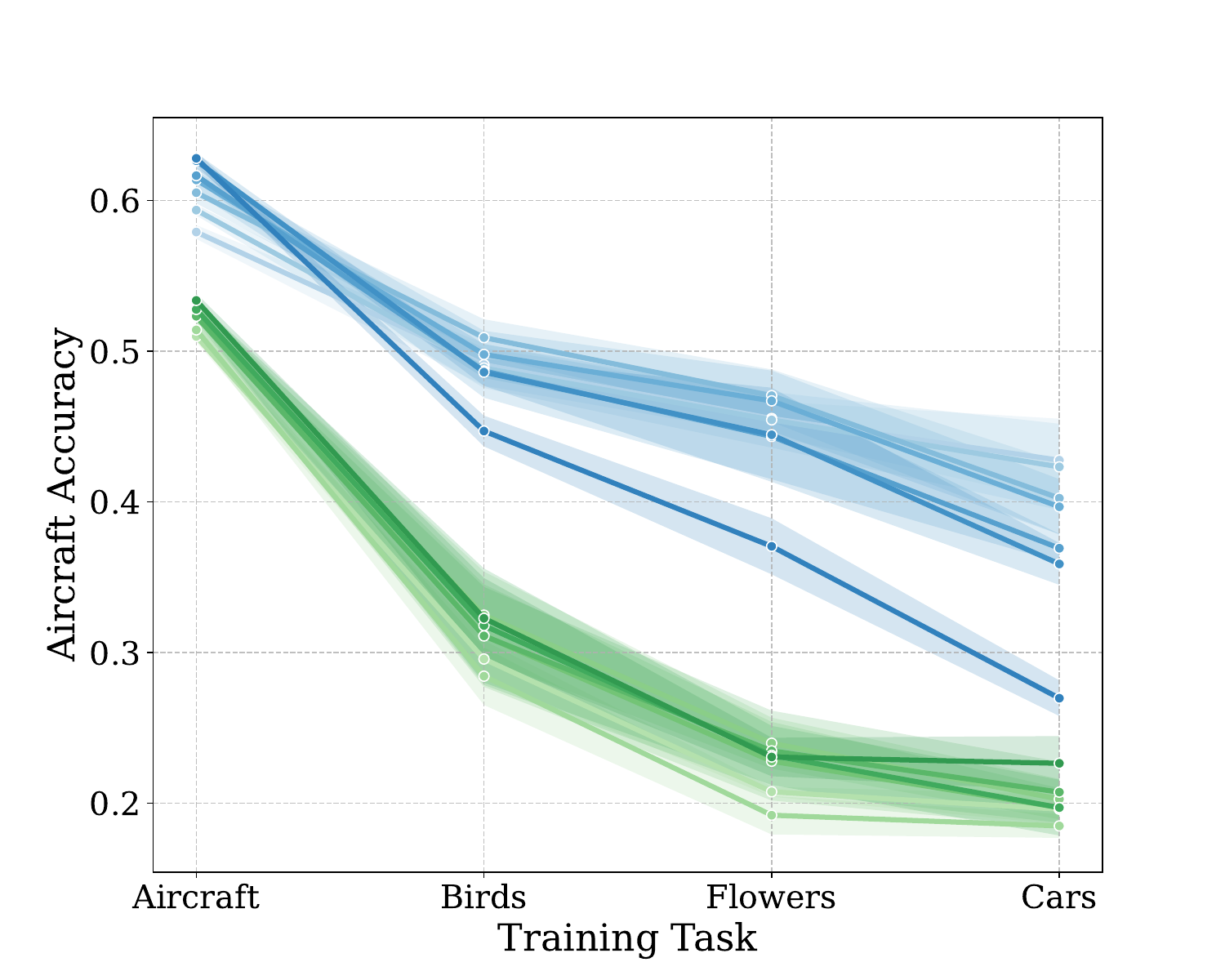}
  \end{subfigure}%
  \begin{subfigure}[b]{0.2\textwidth}
    \includegraphics[trim=400 0 0 0, clip, width=0.85\linewidth]{figures/legend_nofull.pdf}
  \end{subfigure}
  \caption{Test Accuracy on the Imagenet dataset (Left) and on the Aircraft dataset (Right) when training on the \textit{Short Setting} with multiple LoRA ranks, using ViT network (Blue) and ResNet-50 (Green). Results are averaged over 4 runs using different random seed}
  \label{fig:lora_additional_order2}
\end{figure}

\begin{figure}[H]
  \centering
  \begin{subfigure}[b]{0.4\textwidth}
    \includegraphics[trim=10 0 70 50, clip, width=0.95\linewidth]{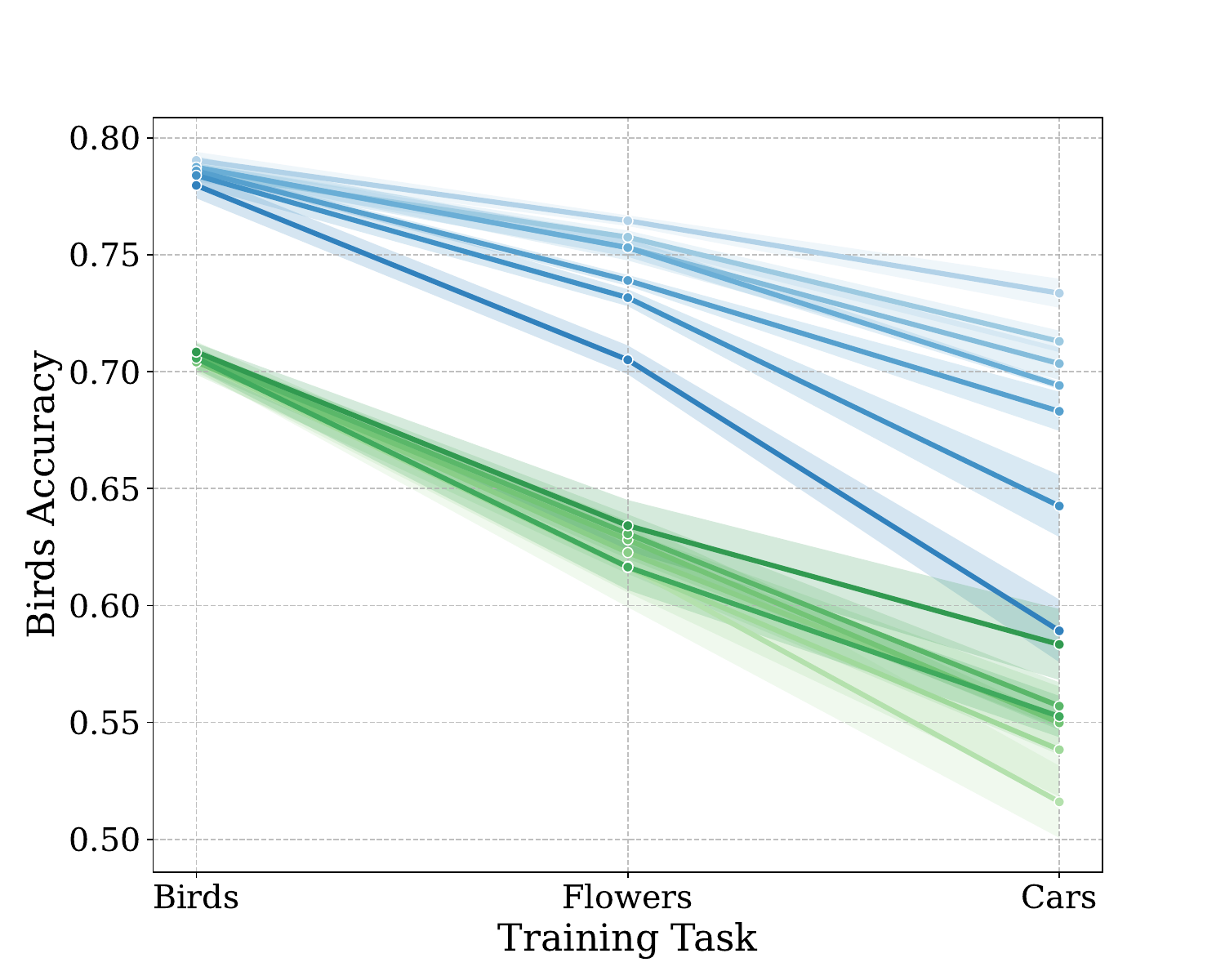}
  \end{subfigure}%
  \begin{subfigure}[b]{0.4\textwidth}
    \includegraphics[trim=10 0 70 50, clip, width=0.95\linewidth]{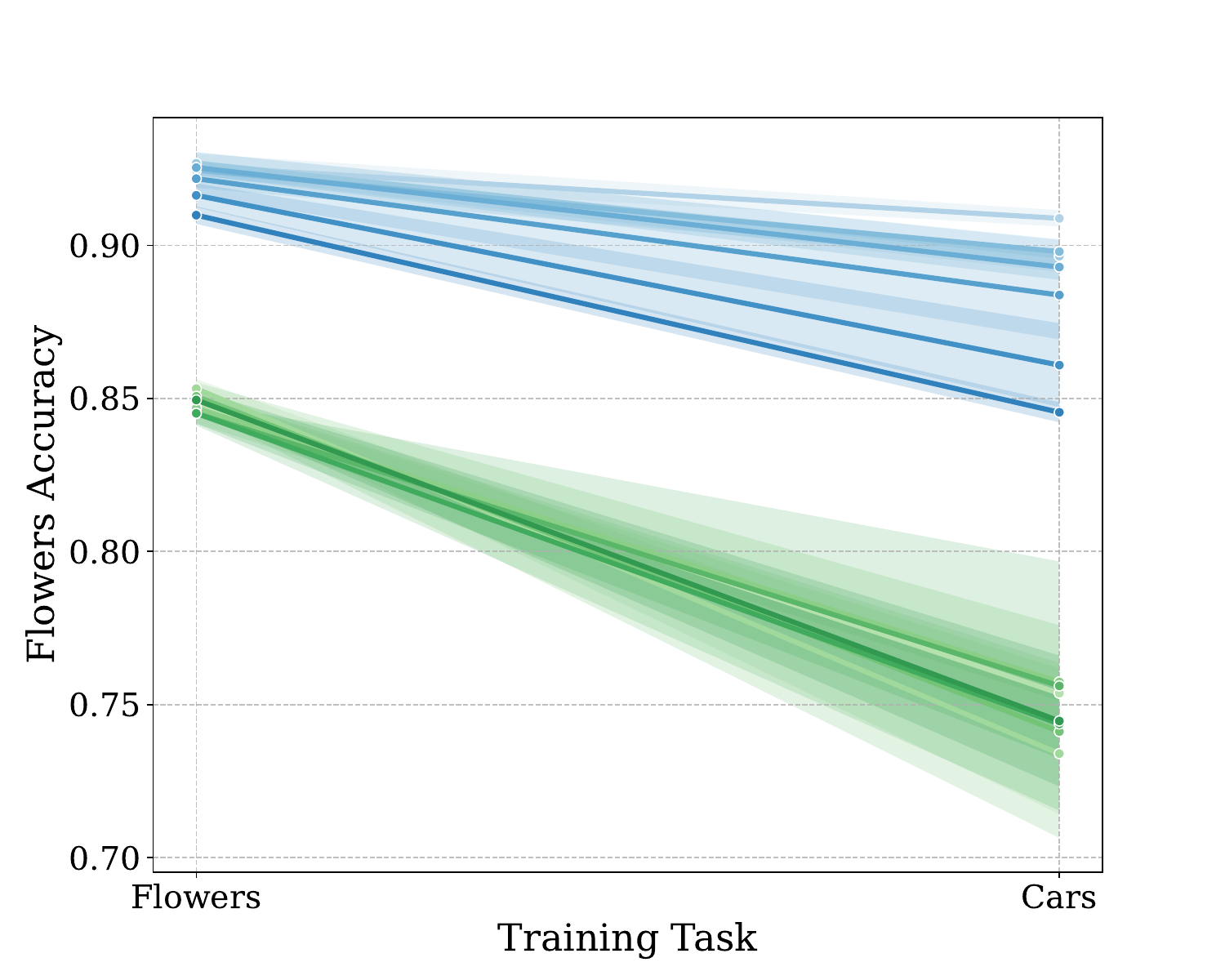}
  \end{subfigure}%
  \begin{subfigure}[b]{0.2\textwidth}
    \includegraphics[trim=400 0 0 0, clip, width=0.85\linewidth]{figures/legend_nofull.pdf}
  \end{subfigure}
  \caption{Test Accuracy on the Birds dataset (Left) and on the Flowers dataset (Right) when training on the \textit{Short Setting} with multiple LoRA ranks, using ViT network (Blue) and ResNet-50 (Green). Results are averaged over 4 runs using different random seed.}
  \label{fig:lora_additional_lwf_order2}
\end{figure}

\begin{table}[H]
    \centering
    \begin{minipage}{0.5\textwidth}
        \centering
        \begin{tabular}{ccc}
            \hline
            \textbf{Rank} & \textbf{Average Accuracy} & \textbf{Average Forgetting} \\
            \hline
            1 & 72.3 $\pm$ 0.5 & 6.5 $\pm$ 0.9 \\
            2 & 71.4 $\pm$ 0.6 & 8.1 $\pm$ 0.8 \\
            4 & 70.6 $\pm$ 0.6 & 9.6 $\pm$ 0.9 \\
            6 & 69.8 $\pm$ 0.4 & 10.9 $\pm$ 0.3 \\
            8 & 68.0 $\pm$ 0.3 & 12.5 $\pm$ 0.3 \\
            16 & 66.0 $\pm$ 0.2 & 15.8 $\pm$ 0.2 \\
            32 & 60.6 $\pm$ 0.4 & 22.3 $\pm$ 0.4 \\
            \hline
        \end{tabular}
    \end{minipage}%
    \begin{minipage}{0.5\textwidth}
        \centering
        \begin{tabular}{ccc}
            \hline
            \textbf{Rank} & \textbf{Average Accuracy} & \textbf{Average Forgetting} \\
            \hline
            1 & 54.4 $\pm$ 0.6 & 21.5 $\pm$ 0.7 \\
            2 & 54.4 $\pm$ 0.3 & 22.1 $\pm$ 0.7 \\
            3 & 55.6 $\pm$ 0.8 & 20.7 $\pm$ 1.7 \\
            4 & 55.4 $\pm$ 0.7 & 21.4 $\pm$ 0.8 \\
            5 & 55.8 $\pm$ 1.6 & 20.7 $\pm$ 2.0 \\
            6 & 55.4 $\pm$ 0.6 & 21.4 $\pm$ 0.8 \\
            7 & 57.0 $\pm$ 0.6 & 19.9 $\pm$ 0.8 \\
            \hline
        \end{tabular}
    \end{minipage}
    \caption{Final average accuracy and forgetting after learning on the Short Setting when learning the tasks in the order Imagenet - Aircraft - Birds - Flowers - Cars, depending on the rank, for ViT (Left) and ResNet (Right). Results are averaged over 4 runs using different seeds.}
    \label{tab:table_forgetting_other_task_order}
\end{table}

\end{document}